\documentclass[journal,compsoc]{IEEEtran}

\IEEEoverridecommandlockouts
\usepackage{cite}
\usepackage{amsmath,amssymb,amsfonts}
\usepackage{algorithmic}
\usepackage{graphicx}
\usepackage{textcomp}
\usepackage{amsfonts}
\usepackage{color}
\usepackage{colortbl}
\usepackage{etoolbox}
\usepackage{multirow}
\usepackage[binary-units=true]{siunitx}
\usepackage[table]{xcolor}
\usepackage{comment}
\usepackage{epsfig}
\usepackage{amsmath}
\usepackage{mathrsfs}
\usepackage{amssymb}
\usepackage[colorlinks]{hyperref}

\definecolor{lower}{RGB}{232, 161, 148}
\definecolor{higher}{RGB}{148, 187, 232}

\newcommand{\ie}{i.e.}
\newcommand{\eg}{e.g.}
\newcommand{\etal}{et al.}
\newcommand{\gt}{ground truth}

\newcommand{\pyd}{PyDNet}
\newcommand{\fastdepth}{FastDepth}
\newcommand{\resnet}{MonoDepth2}

\newcommand{\dsnet}{DSNet}
\newcommand\norm[1]{\left\lVert#1\right\rVert}

\begin{document}

\title{Real-time single image depth perception\\in the wild with handheld devices}
\author{Filippo~Aleotti,
        Giulio~Zaccaroni,
        Luca~Bartolomei, 
        Matteo~Poggi,
        Fabio~Tosi,
        Stefano Mattoccia
}

\IEEEtitleabstractindextext{%
\begin{abstract}
Depth perception is paramount to tackle real-world problems, ranging from autonomous driving to consumer applications. For the latter, depth estimation from a single image represents the most versatile solution, since a standard camera is available on almost any handheld device. Nonetheless, two main issues limit its practical deployment: i) the low reliability when deployed in-the-wild and ii) the demanding resource requirements to achieve real-time performance, often not compatible with such devices.
Therefore, in this paper, we deeply investigate these issues showing how they are both addressable adopting appropriate network design and training strategies -- also outlining how to map the resulting networks on handheld devices to achieve real-time performance. Our thorough evaluation highlights the ability of such fast networks to generalize well to new environments, a crucial feature required to tackle the extremely varied contexts faced in real applications. Indeed, to further support this evidence, we report experimental results concerning real-time depth-aware augmented reality and image blurring with smartphones in-the-wild.
\end{abstract}

}

\maketitle

\IEEEpeerreviewmaketitle

\section{Introduction}

Depth perception is an essential step to tackle real-world problems such as robotics and autonomous driving, and some well-known sensors exist for this purpose. Among them, active sensing techniques such as Time-of-Flight (ToF) or LiDAR are often deployed in the application domains mentioned before. However, they struggle with typical consumer applications since ToF is mostly suited for indoor environments. At the same time, conventional LiDAR technology, frequently used for autonomous driving and other tasks, is too cumbersome and expensive for its deployment with relatively cheap and lightweight consumer handheld devices. Nonetheless, it is worth noting that active sensing technologies, mostly suited for indoor environments, are sometimes integrated into high-end devices as, for instance, occurs with the 2020 Apple iPad Pro.

Therefore, camera-based technologies are often the only viable strategies to infer depth in consumer applications, and among them, well-known methodologies are structured-light and stereo vision. The former typically requires an infrared camera and a specific pattern projector making it not suited for environments flooded by sunlight. The latter requires two appropriately spaced and synchronized cameras. Some recent high-end smartphones or tablets feature one or both technologies, although they are not yet widespread enough to be considered as standard equipment. Moreover, for stereo systems, the distance between the cameras (baseline) is necessarily narrow, limiting the depth range to a few meters away.

\begin{figure}[t]
	\centering
	\includegraphics[width=1.0\linewidth]{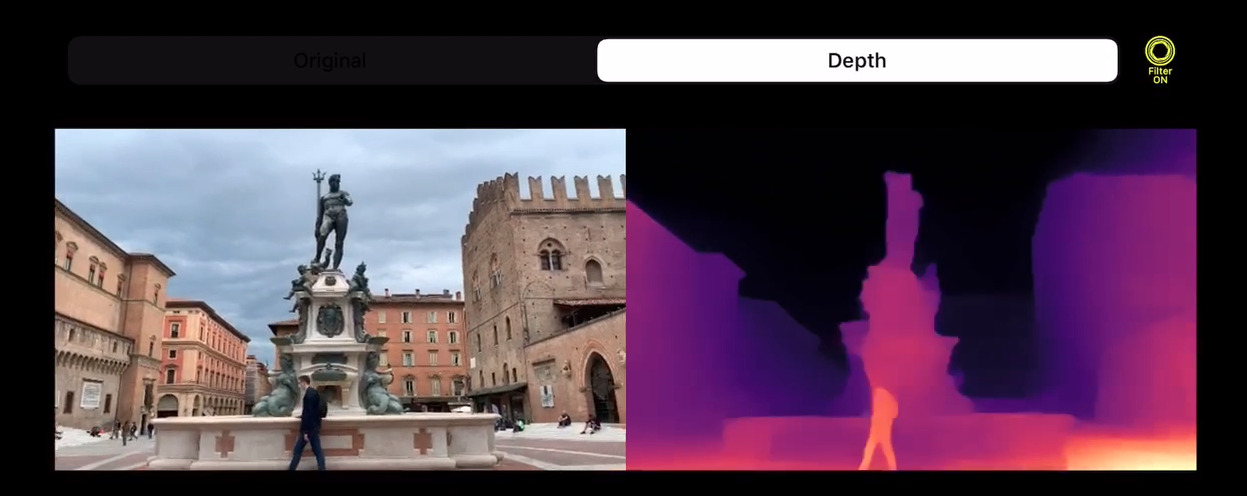}
	\caption{\textbf{Depth perception in the wild with a mobile app.} Single image depth perception in the wild at nearly 60 FPS with an iPhone XS and the \pyd{} \cite{pydnet18} network.}
	\label{fig:teaser}
\end{figure}

On the other hand, with the advent of deep-learning, recent years witnessed the rising of a further strategy to infer depth using a single standard camera available substantially in any consumer device. Compared to previous technologies for depth estimation, such an approach would enable to tackle all the limitations mentioned before. Nonetheless, single image depth estimation is seldom deployed in consumer applications due to two main reasons. The first one concerns the low reliability of state-of-the-art methods when tackling depth estimation in the wild dealing with unpredictable environments not seen at training time, as necessarily occurs when targeting a massive amount of users facing heterogeneous environments. The second reason concerns the constrained computational resources available in handheld devices, such as smartphones or tablets, deployed for consumer applications. In fact, despite the steady progress in this field, the gap with system leveraging high-end GPUs is (and always will be) significant since power requirements heavily constrain handheld devices. Despite the limited computational resources available, most consumer applications need real-time performance.      

Arguing these facts, in this paper, we deeply investigate both the issues outlined so far. In particular, we will show how tackling them leveraging appropriate network design approaches, training strategies, and outlining how to map them on off-the-shelf consumer devices to achieve real-time performance, as shown in Figure \ref{fig:teaser}. Indeed, our extensive evaluation highlights the ability of the resulting networks to robustly generalize to unseen environments, a crucial feature to tackle the heterogeneous contexts faced in real consumer applications.

The paper is organized as follows. At first, we present previous works about monocular depth estimation on high performance and mobile devices, then we describe the framework that allows us to train a model, even a lightweight one, to be robust when deployed in the wild. We then evaluate a set of monocular networks already proposed in the literature on three benchmarks, deploying such models on mobile smartphones. Finally, we report how enabling real-time single-image depth estimation at the edge can be effectively exploited to tackle two well-known applications: depth aware augmented reality and blurring.

\section{Related work}
\textbf{Monocular depth estimation.} Even if depth estimation from multiple views has a long history in computer vision, depth from a single image \cite{saxena2009make3d,ladicky2014pulling} started being considered feasible only with the advent of deep learning techniques. Indeed, obtaining depth values starting from a single image is an \textit{ill-posed} problem, since an infinite number of real-world layouts may have generated the input image. However, learning-based methods, and in particular deep learning, proved to be adequate to face the problem.  Initially were proposed supervised approaches \cite{Eigen_2014,Laina_3DV_2016}. However, the need for \gt{} measurements represents a severe restraint since an active sensor, as a LiDAR, together with specific manual post-processing is required to obtain such data.  Therefore, effective deployment of supervised approaches is burdensome both in terms of time and costs. To overcome this limitation, solutions able to learn depth supervised only by images are incredibly appealing, and nowadays, several methods rely on simple stereo pairs or monocular video sequences for training.
Among these, \cite{Godard_CVPR_2017} is the first notable attempt leveraging stereo pairs, eventually improved exploiting traditional stereo algorithm \cite{Tosi_2019_CVPR, Watson_ICCV_2019}, visual odometry supervision \cite{Yang_ECCV_2018, Andraghetti_3DV_2019} or 3D movies \cite{midas}. On the other hand, methods leveraging monocular videos do not even require a stereo camera at training time, at the cost of learning depth estimation up to a scale factor. Starting with Zhou \etal{} \cite{Zhou_2017_CVPR}, proposing to learn both depth and camera ego-motion, more recent methods propose to apply Direct Visual Odometry \cite{Wang_CVPR_2018} or ICP \cite{Mahjourian_CVPR_2018} strategies to improve predictions. Additional cues have been used by more recent works, such as optical flow \cite{Yin_CVPR_2018,Zou_ECCV_2018,Chen_ICCV_2019, Ranjan_CVPR_2019} or semantic segmentation \cite{omeganet}. Finally, it is worth noting that the two supervisions coming from stereo images and monocular sequences can be combined \cite{Godard_ICCV_2019}.

\textbf{Depth estimation on mobile systems.} Mobile devices are ubiquitous, and deep learning opened many applications scenarios \cite{howard2017mobilenets}. Although sometimes server-side inference is unavoidable, maintaining the computation on-board is highly beneficial. It allows to get-rid of privacy issues, the need for tailored datacenters allowing to reduce costs as well to improve scalability. Moreover, although not critical for most consumer scenarios, full on-board processing does not dictate an internet connection.  
Despite limited computing capabilities, mostly constrained by power consumption issues and typically not comparable to those available in standard laptops or PCs, some authors proposed deep networks for depth estimation suited for mobile devices too. These works targeted stereo \cite{wang2018anytime} and monocular \cite{pydnet18, icra_2019_fastdepth} setups. Moreover, some authors proposed depth estimation architectures tailored for specific hardware setups, such as those based on dual-pixel sensors available in some recent Google's smartphones, as reported in \cite{garg2019learning,zhang20202}.

\section{Problems and requirements}
The availability of more and more powerful devices paves the way for complex and immersive applications, in which users can interact with the nearby environment. As a notable example, augmented reality can be used to display interactive tools or concepts, avoiding to build a real prototype and thus cutting off costs. For this and many other applications, obtaining accurate depth information with a high frame rate is paramount to further enhance the interaction with the surrounding environment, even with devices devoid of active sensors.  
Almost any modern handheld device features at least a single camera and an integrated CPU within, typically, an ARM-based system-on-chip to cope with the constrained energy budget of such devices. Sometimes, especially in most new ones, a Neural Processing Unit (NPU) devoted to accelerating deep neural networks is also available. Inevitably, the resulting overall computation performance is far from conventional PC-based setups, and the availability of an NPU only partially fills this gap. Given these constraints, single-image depth perception would be rather appealing since it could seamlessly deal with dynamic contexts, whereas other techniques such as \emph{structure from motion} (SfM) would struggle. However, these techniques are computationally demanding, and most state of the art approaches would not fit the computational resources available in handheld devices. Moreover, regardless of the computing requirements, training the networks for predictable target environments is not feasible for consumer applications. Thus the depth estimation network shall be robust to any faced deployment scenarios and possibly invariant to the training data distribution. 
A client-server approach would soften some of the computational issues, although with notable disadvantages — the need for an internet connection and a poorly scaling of the whole overall system when the number of users increases.      

To get rid of all the issues mentioned above and to deal with practical applications, we will describe next how to achieve real-time and robust single image depth perception on low-power architectures found in off-the-shelf handheld devices.

\section{Framework overview}
In this section, we introduce our framework aimed at enabling single image depth estimation in the wild with mobile devices, devoting specific attention to iOS and Android systems. Before the actual deployment on the target handheld device, our strategy requires an offline training procedure typically carried out on power unconstrained devices such as a PC equipped with a high-end GPU. We will discuss in the reminder the training methodology, leveraging \textit{knowledge distillation}, deployed to achieve our goal in a limited amount of time and the dataset adopted for this purpose.  
Another critical component of our framework is a lightweight network enabling real-time processing on the target handheld devices. Purposely, we will introduce and thoroughly assess the performance of state of the art networks fitting this constraint.

\subsection{Off-line training} 
As for most learning-based monocular depth estimation models, our proposal is trained off-line on standard workstations, equipped with one or more GPUs, or through cloud processing services. In principle, depending on the training data available, one can leverage different training strategies: supervised, semi-supervised or self-supervised training paradigms. Moreover, as done in this paper, cheaper and better-scaling supervision can be conveniently obtained from another network, by leveraging knowledge distillation to avoid the need for expensive \gt{} labels, through a teacher-student network. 

When a large enough dataset providing \gt{} labels inferred by an active sensor is available, such as \cite{Matterport3D, diode_dataset}, (semi-)supervised fashion is certainly valuable since it enables, among other things,  to disambiguate difficult regions (\eg{} texture-less regions such as walls). Unfortunately, large datasets with depth labels are not available or extremely costly and cumbersome to obtain. Therefore, when this condition is not met, self-supervised paradigms enable to train with (potentially) countless examples, at the cost of a more challenging training setup and typically less accurate results. Note that, depending on the dataset, a strong depth prior can be distilled even if are not available depth labels provided by an active sensor. For instance, \cite{Tosi_2019_CVPR, Watson_ICCV_2019} exploit depth values from a stereo algorithm, while \cite{li2019learning} relies on a SfM pipeline. Finally, supervision can be distilled from other networks as well, for the stereo \cite{guo2018learning} and monocular \cite{Poggi_CVPR_2020} setup. The latter is the strategy followed in this paper. Specifically, we use as teacher the \textit{MiDaS} network proposed in \cite{midas}. 
This strategy allows us to speed-up the training procedure of the considered lightweight networks significantly, since doing this from scratch according to the methodology proposed in \cite{midas} would take much much longer time (weeks vs days) since mostly bounded by proxy labels generation.
Moreover, it is worth noting that given a reliable teacher network, pre-trained in a semi or self-supervised manner, such as \cite{midas}, it is straightforward to distill an appropriate training dataset since any collection of images is potentially suited to this aim. We will describe next the training dataset used for our experiments made of a bunch of single images belonging to well-known popular datasets.

\subsection{On-device deployment and inference} 

Once outlined the training paradigm, the next issue concerns the choice of a network capable of learning from the teacher how to infer meaningful depth maps and, at the same time, able to run in real-time on the target handheld devices. Unfortunately, only a few networks described next potentially fulfil these requirements, in particular, considering the ability to run in real-time on embedded systems.  
 
Once identified and trained a suitable network, its mapping on a mobile device is nowadays quite easy. In fact, there exist various tools that, starting from a deep learning framework as PyTorch \cite{paszke2017automatic} or TensorFlow \cite{tensorflow2015-whitepaper}, can export, optimize (\eg{} perform weights quantization) and execute models even leveraging mobile GPU \cite{lee2019device} on principal operating systems (OS). In some cases, the target OS exposes utilities and tools to improve the performances further. For instance, starting from iOS 13, neural networks deployed on iPhones can use the GPU or even the Apple Neural Engine (ANE) thanks to Metal and Metal Performance Shaders (MPS), thus largely improving the runtime performances. We will discuss in the next section how to map the networks on iOS and Android devices using  TensorFlow and PyTorch as high-level development frameworks.

\section{Lightweight networks for single image depth estimation}

According to the previous discussion, only a subset of the state of the art single image depth estimation networks fits our purposes. Specifically, we consider the following publicly available lightweight architectures: \pyd{} \cite{pydnet18}, \dsnet{} \cite{omeganet} and \fastdepth{} \cite{icra_2019_fastdepth}. Moreover, we also include a representative example of a large state of the art network \resnet{}, proposed in \cite{Godard_ICCV_2019}. It is worth to notice that other and more complex state-of-the-art networks, as \cite{Tosi_2019_CVPR}, could be deployed in place within the proposed framework. However, this might come at the cost of higher execution time on the embedded device and, potentially, overhead for the developer in case of custom layers not directly supported by the mobile executor (\eg, the correlation layer used in \cite{Tosi_2019_CVPR}).

\textbf{\resnet.} An architecture deploying a ResNet encoder, proposed initially in \cite{he2016deep}, made of 18 feature extraction layers, shrinking the input by a factor of $\frac{1}{32}$. Then, the dense layers are replaced in favour of a decoder module, able to restore the original input resolution and output an estimated depth map. At each level in the decoder, $3\times3$ convolutions with skip connections are performed, followed by a $3\times3$ convolution layer in charge of depth estimation. The resulting network can predict depths at different scales, counting 14.84 M parameters. It is worth to notice that in our evaluation we do not rely on ImageNet \cite{deng2009imagenet} pre-training for the encoder for fairness to other architectures not pre-trained at all.

\textbf{\pyd.} This network, proposed in \cite{pydnet18}, features a pyramidal encoder-decoder design able to infer depth maps from a single RGB image. Thanks to its small size and design choices, \pyd{} can run on almost any device including low-power embedded platforms \cite{DATE_2019}, such as the Raspberry Pi 3. In particular, the network exploits 6 layers to reduce the input resolution at $\frac{1}{64}$, restored in the depth domain by 5 layers in the decoder. Each layer in the decoder applies $3\times3$ convolutions with $96,64,32,8$ feature channels, followed by a $3\times3$ convolution in charge of depth estimation.
Notice that, to keep low the resources and inference time, top prediction of \pyd{} is at half resolution, so the final depth map is obtained through an upsampling operation. We adopt the mobile implementation provided by the authors, publicly available online\footnote{\url{https://github.com/FilippoAleotti/mobilePydnet}}, which differs from the paper network by small changes (\eg{} transposed convolutions have been replaced by upsampling and convolution blocks). The network counts 1.97 M parameters.

\textbf{\fastdepth.} Proposed by Wofk \etal{} \cite{icra_2019_fastdepth}, this network can infer depth predictions at 178 fps with an NVIDIA Jetson TX2 GPU. This notable speed is the result of design choices and optimization steps. Specifically,  the encoder is a MobileNet \cite{howard2017mobilenets}, thus suited for execution on embedded devices. The decoder consists of 6 layers, each one with a depth-wise separable convolution, with skip connections starting from the encoder (in this case, features are combined with addition). However, it is worth observing that the highest frame rate previously reported is achievable only exploiting both pruning \cite{yang2018netadapt} and hardware-specific optimization techniques. In this paper, we do not rely on such strategies for fairness with other networks. The network counts 3.93 M parameters.

\textbf{\dsnet} This architecture is part of $\Omega$Net \cite{omeganet}, an ensemble of networks predicting not only the depth of the scene starting from a single view but also the semantic segmentation, camera intrinsic parameters and if two frames are provided, the optical flow. In our evaluation we consider only the depth estimation network \dsnet{}, inspired by \pyd{}, which contains a feature extractor able to decrease the resolution by $\frac{1}{32}$, followed by 5 decoding layers able to infer depth predictions starting from the current features and previous depth estimate. In the original architecture, the last decoder also predicts per-pixel semantic labels through a dedicated layer, removed in this work. With this change, the network counts 1.91 M of parameters, 0.2 M fewer than the original model.

\section{Datasets}

In our evaluation, we use four datasets. At first, we rely on the KITTI dataset to assess the performance of the four networks when trained with the standard self-supervised paradigm deployed typically in this field \cite{Godard_ICCV_2019}. Then, we re-train from scratch the four networks using the paradigm previously outlined, distilling proxy labels by employing the pre-trained MiDaS network \cite{midas} made available by the same authors. For this task, we use a novel dataset, referred to as WILD, described next. We then evaluate the networks trained according to this methodology on the TUM RGBD \cite{sturm12iros} and NYUv2 \cite{Silberman:ECCV12} dataset to assess their generalization capability. 

\textbf{KITTI.} The KITTI dataset \cite{Menze2015CVPR} contains 61 scenes collected by moving car equipped with a LiDAR sensor and a stereo rig. Following \cite{Godard_ICCV_2019}, we select a split of 697 images for testing, while 39810 and 4424 images are used respectively for preliminary training and validation purpose. Moreover, we use it to assess the generalization capability of the networks in the wild during the second part of our evaluation. 

\textbf{WILD} The Wild dataset (W), introduced in this paper, consists of a mixture of Microsoft COCO \cite{lin2014microsoft} and OpenImages \cite{OpenImages2} datasets. Both datasets contain a large number of internet photos, and they do not provide depth labels. Moreover, since video sequences nor stereo pairs are available, they are not suited for conventional self-supervised guidance methods (\eg{} SfM or stereo algorithms). On the other hand, they cover a broad spectrum of various real-world situations, allowing to face both indoor and outdoor environments, deal with everyday objects and various depth ranges. We select almost 447000 frames for training purposes\footnote{Details concerning the WILD dataset are available at this link: \\ \url{https://github.com/FilippoAleotti/mobilePydnet}}. Then, we distilled the supervision required by our networks with the robust monocular architecture proposed in \cite{midas} with the weights publicly available.
We point out once again that our supervision protocol has been carefully chosen mostly for practical reasons. It takes a few days to distill the WILD dataset by running MiDaS (using the publicly available checkpoints) on a single machine. On the contrary, to obtain the same data used to train the network as in \cite{midas}, it would require an extremely intensive effort. Doing so, we can scale better: since we trust the teacher, we could, in principle, source knowledge from various and heterogeneous domains on the fly. Of course, the major drawbacks of this approach are evident:  we need an already available and reliable teacher, and the accuracy of the student is bounded to the one of the teacher. However, we point out that the training scheme proposed in \cite{midas} is general, so it can also be applied in our case, and that we already expect a margin with state-of-the-art networks due to the lightweight size of mobile architectures considered. For these reasons, we believe that our approach is beneficial to source a fast prototype than can be improved later leveraging other techniques if needed. This belief is supported by experimental results presented later in the paper.    

\textbf{TUM RGBD.} The TUM RGBD (3D Object Reconstruction category) dataset \cite{sturm12iros} contains indoor sequences framing people and furniture. We adopt the same split of 1815 images used in \cite{li2019learning} for evaluation purposes only.

\textbf{NYUv2.} The NYUv2 dataset \cite{Silberman:ECCV12} is an indoor RGBD dataset acquired with a Microsoft Kinect device. It provides more than 400k raw depth frames and 1449 densely labelled frames. As for the previous dataset, we adopt the official test split containing 649 images for generalization tests.

\section{Mapping on mobile devices}

Since we aim at mapping single image depth estimation networks on handheld devices, we briefly outline here the steps required to carry out this task.

\begin{figure}[t]
     \renewcommand{\tabcolsep}{1.2px}
    \centering
    \begin{tabular}{cc}
    \includegraphics[width=0.5\linewidth]{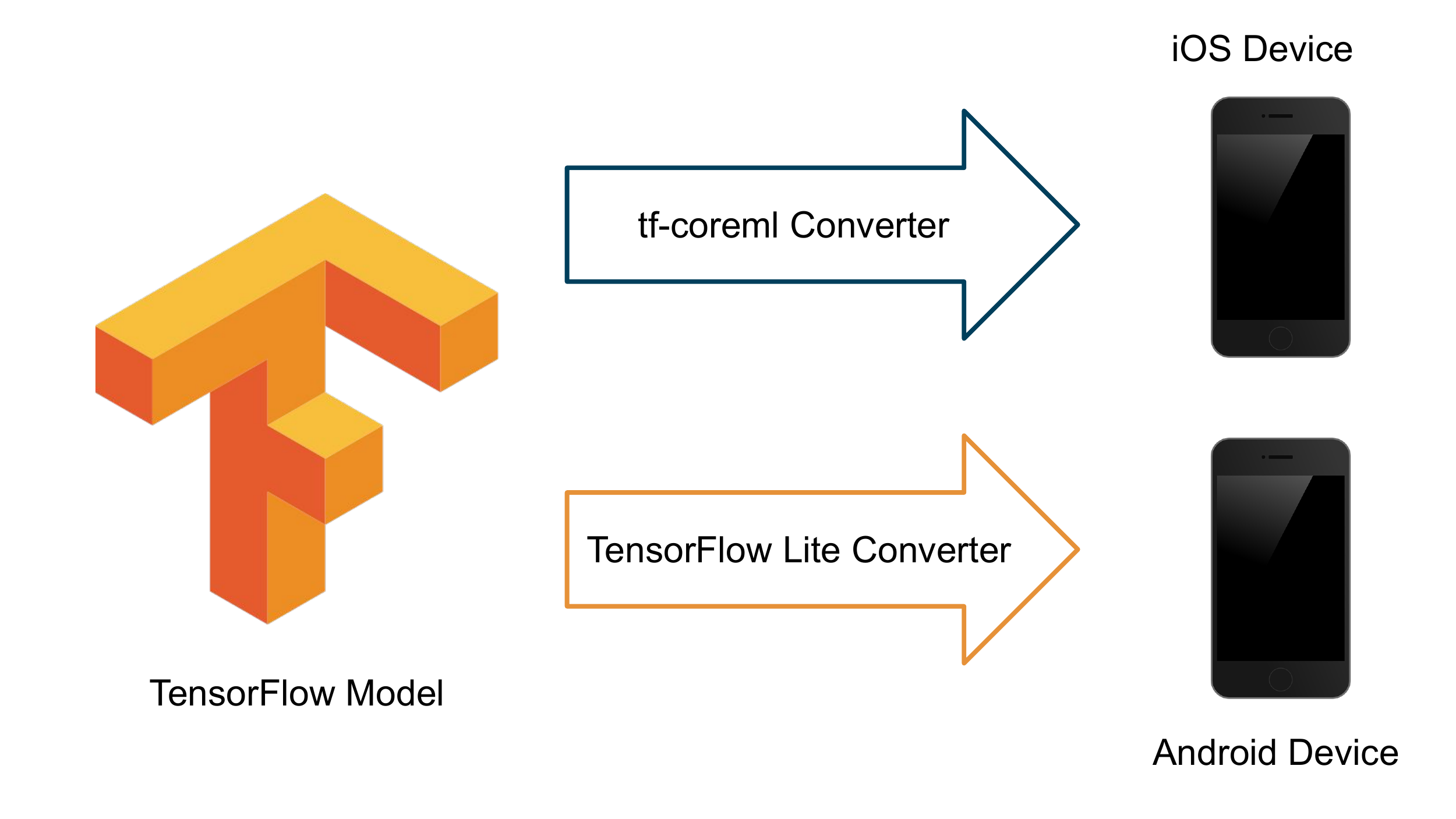} &
    \includegraphics[width=0.5\linewidth]{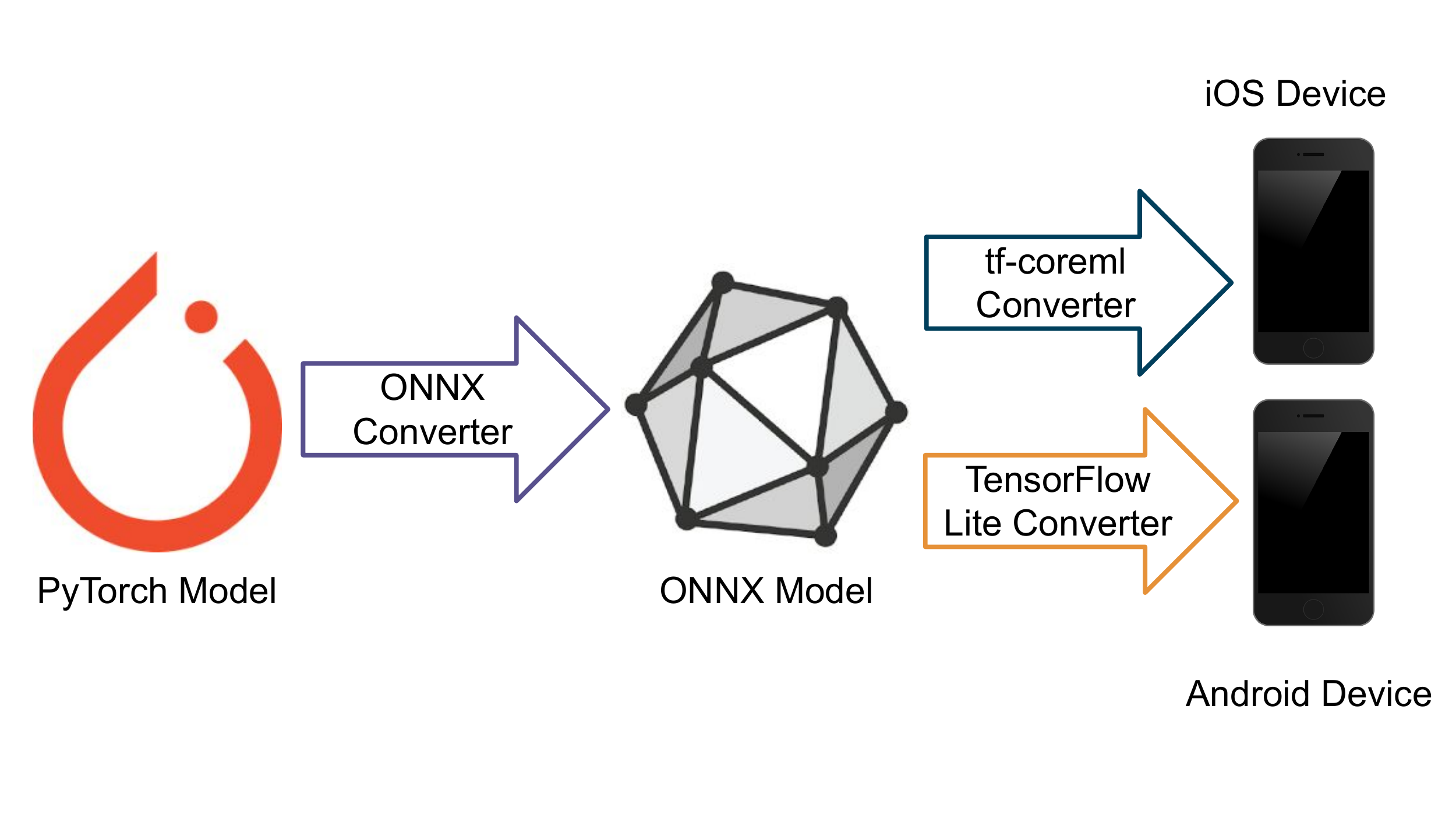} \\
    TensorFlow to mobile & PyTorch to mobile
    \end{tabular}
    \caption{\textbf{Porting a network from a standard ML frameworks to a mobile device.} Both TensorFlow and PyTorch models can be easily converted into models suited for mobile execution using libraries available for each OS.}
    \label{fig:pipelines}
\end{figure}

\begin{table*}[t]
\centering
\setlength{\tabcolsep}{6pt} 
\begin{tabular}{l|cccc|ccc}
\multicolumn{3}{c}{} & \multicolumn{2}{c}{\cellcolor{lower} Lower is better} & \multicolumn{2}{c}{\cellcolor{higher} Higher is better}\\
\hline
Network  &\cellcolor{lower}Abs Rel & \cellcolor{lower}Sq Rel & \cellcolor{lower}RMSE & \cellcolor{lower}log RMSE & \cellcolor{higher}$\delta<$1.25 &  \cellcolor{higher}$\delta<1.25^2$ & \cellcolor{higher}$\delta<1.25^3$  \\
\hline
\pyd{}   & 0.153  &  1.363  &  6.030 & 0.252 &  0.789 &  0.918  & 0.963 \\ 
\fastdepth{}  & - & - & - & - & - &  - & - \\ 
\dsnet{}   & 0.130& 0.909 & 5.022 & 0.207 & 0.842 & 0.948 & 0.979 \\
\hline
\resnet{} \textdagger & \textbf{0.132} & \textbf{1.044} & \textbf{5.142} & \textbf{0.210} & \textbf{0.845} & \textbf{0.948} & \textbf{0.977} \\
\pyd \textdagger & 0.154  &   1.307  &   5.556  &   0.229  &   0.812  &   0.932  &   0.970 \\ 
\fastdepth \textdagger &  0.156  &   1.260  &   5.628  &   0.231  &   0.801  &   0.930  &   0.971   \\
\dsnet \textdagger &   0.159  &   1.272  &   5.593  &   0.233  &   0.800  &   0.932  &   0.971  \\
\hline
\end{tabular}
\caption{\textbf{Quantitative results on Eigen split.} \textdagger{} indicates models trained according to \cite{Godard_ICCV_2019} training framework, otherwise we report results provided in each original paper.}
\label{table:eigen}
\end{table*}

\begin{figure*}[t]
    \centering
    \renewcommand{\tabcolsep}{1.0px}
    \begin{tabular}{ccccc}
    \includegraphics[width=0.2\linewidth]{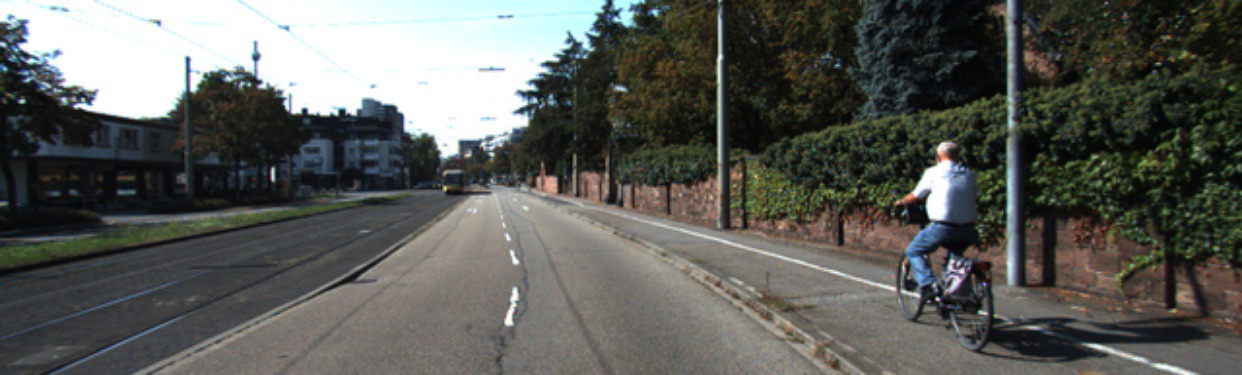} &
    \includegraphics[width=0.2\linewidth]{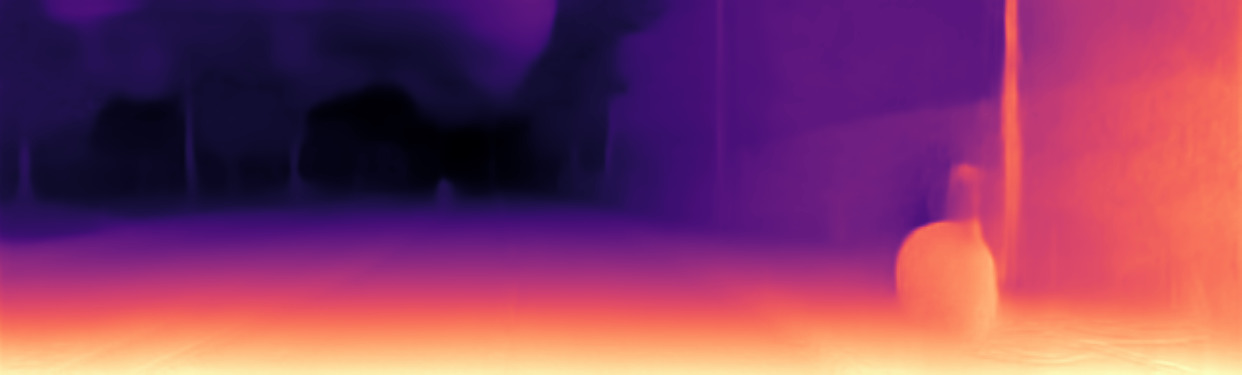} &
    
    \includegraphics[width=0.2\linewidth]{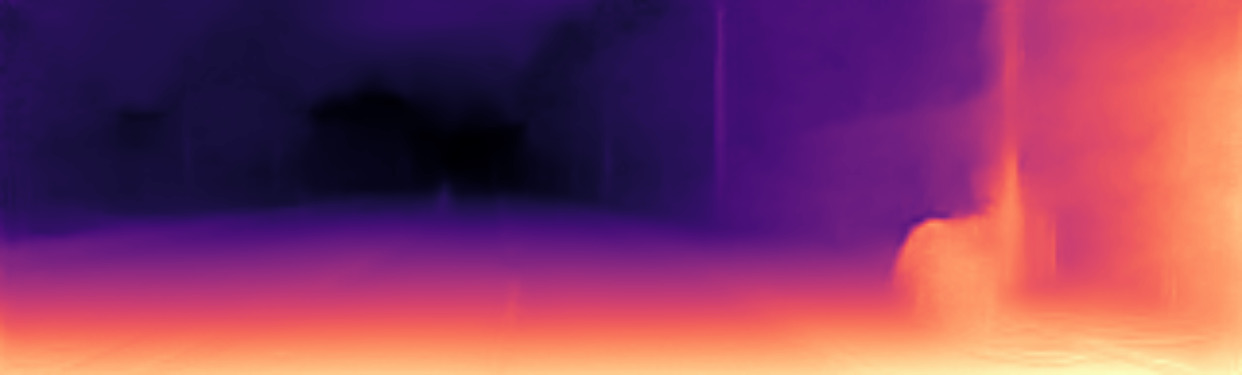} &
    \includegraphics[width=0.2\linewidth]{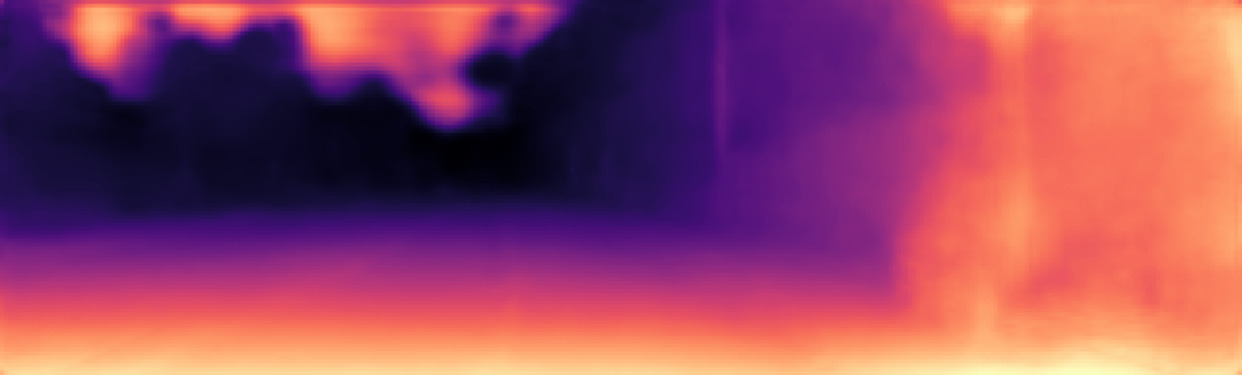} &
    \includegraphics[width=0.2\linewidth]{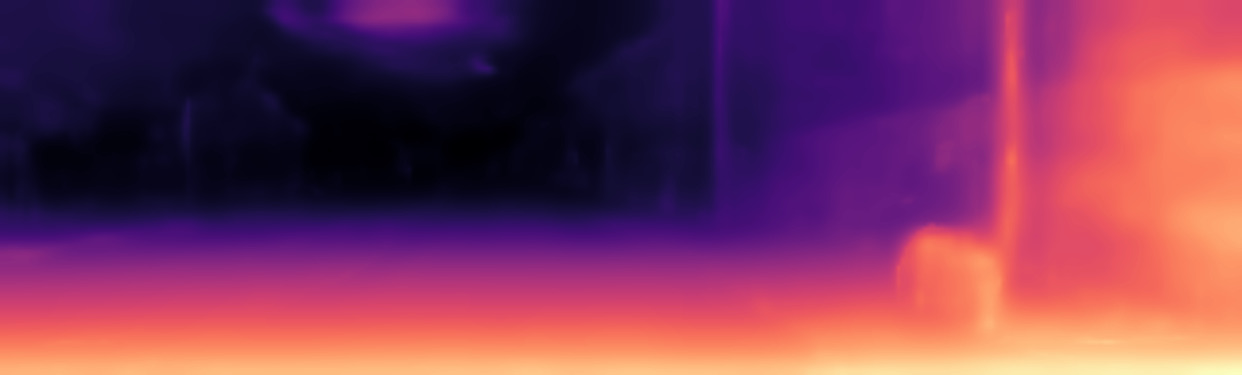} \\
    Reference image & \resnet{} & \pyd{} & \fastdepth{} & \dsnet{} \\
    \end{tabular}
    \caption{\textbf{Qualitative results on KITTI.} All the models have been trained equally using the framework of \cite{Godard_ICCV_2019} on the Eigen split of KITTI.}
    \label{fig:eigen}
\end{figure*}

As depicted in Figure \ref{fig:pipelines}, different tools are available according to both the deep learning framework and the target OS. With TensorFlow models, weights are processed using \textit{tf-coreml} converter in case of iOS deployment or TensorFlow \textit{Lite} converter when targeting an Android device.
On the other hand, when starting from PyTorch models, a further intermediate step is required. In particular, stored weights are converted in the Open Neural Network Exchange format (ONNX), a common representation allowing for porting architectures implemented in a source framework into a different target framework. This conversion is possible seamlessly if the networks consist of standard layers, while is not straightforward in case of custom modules (\eg{}, correlation layers as in monoResMatch \cite{Tosi_2019_CVPR}). 
Although these tools typically enable to perform weights quantization during the model conversion phase to the target environment, we refrain from applying quantization to maintain the original accuracy of each network. In the experimental results section, we will provide execution time mapping the networks on mobile devices following the porting strategy outlined so far.

\section{Experimental results}

In this section, we thoroughly assess the performance of the considered networks with standard datasets deployed in this field. At first, since differently from other methods \fastdepth{} \cite{icra_2019_fastdepth} was not initially evaluated on KITTI, we carry out a preliminary evaluation of all networks on such dataset. Then, we train from scratch the considered networks according to the framework outlined on the Wild dataset, evaluating their generalization ability. Finally, we show how to take advantage of the depth maps inferred by such networks for two applications particularly relevant for mobile devices.    

\subsection{Evaluation on KITTI}

At first, we first investigate the accuracy of the considered networks on the KITTI dataset. Since the models have been developed with different frameworks (\pyd{} in TensorFlow, the other two in PyTorch) and trained on different datasets (\fastdepth{} on NYU v2 \cite{Silberman:ECCV12}, others on the Eigen \cite{Eigen_2014} split KITTI \cite{Menze2015CVPR}), we implement all the networks in PyTorch. This strategy allows us to adopt the same self-supervised protocol proposed in \cite{Godard_ICCV_2019} to train all the models. This choice is suited for the KITTI dataset since it exploits stereo sequences enabling to achieve the best accuracy. Given two images $I$ and $I^\dagger$, with known intrinsic parameters ($K$ and $K^\dagger$) and relative pose of the cameras ($R$,$T$), the network predicts depth $\mathcal{D}$ allowing to reconstruct the reference image $I$ from $I^\dagger$, so:

\begin{equation}\label{equation:warping}
    \hat{I} = \omega(I^\dagger, K^\dagger, R, T, K, \mathcal{D})
\end{equation}
where $\omega$ is a differentiable warping function.

Then, the difference between $\hat{I}$ and $I$ can be used to supervise the network, thus improving $\mathcal{D}$, without any \gt{}. The loss function used in \cite{Godard_ICCV_2019} is composed by a photometric error term $p_e$ \ref{equation:pe} and an edge-aware regularization term $L_s$. 

\begin{equation}\label{equation:pe}
    p_e(I, \hat{I}) = \alpha\frac{(1- \mathrm{SSIM}(I, \hat{I}))}{2} + (1-\alpha) \norm{I -\hat{I}}_1
\end{equation}

\begin{equation}
    \mathcal{L}_s = \norm{\delta_x\mathcal{D}_t^*}_1e^{-\norm{\delta_xI_t}_1} + \norm{\delta_y\mathcal{D}_t^*}_1e^{-\norm{\delta_yI_t}_1}
\end{equation}

where $\mathrm{SSIM}$ is the structure similarity index \cite{wang2004image}, while $\mathcal{D*}=\mathcal{D}/\overline{\mathcal{D}}$ is the mean normalized inverse depth proposed in \cite{wang2018learning}. We adopt the M configuration of \cite{Godard_ICCV_2019} to train all the models. Doing so, given the reference image $I_t$, at training time we also need $\{I_{t-1},I_{t+1}\}$, that are respectively the previous and the next frames in the sequence, to leverage the supervision from monocular sequences as well. Purposely, a pose network is trained to estimate relative poses between the frames in the sequence as in \cite{Godard_ICCV_2019}. Moreover, \textit{per-pixel minimum} and \textit{automask} strategies are used to preserve sharp details: the former select best $p_e$ among multiple views according to occlusions, while the latter helps to filter out pixels that do not change between frames (\eg{} scenes with a non-moving camera or dynamic objects that are moving at the same speed of the camera), thus breaking the \textit{moving camera in a stationary world} assumption (more details are provided in the original paper \cite{Godard_ICCV_2019}). Finally, intermediate predictions, when available, are upsampled and optimized at input resolution. 

\begin{figure*}[t]
    \centering
    \renewcommand{\tabcolsep}{1.0px}
    \begin{tabular}{cccc}
    \includegraphics[width=0.25\linewidth]{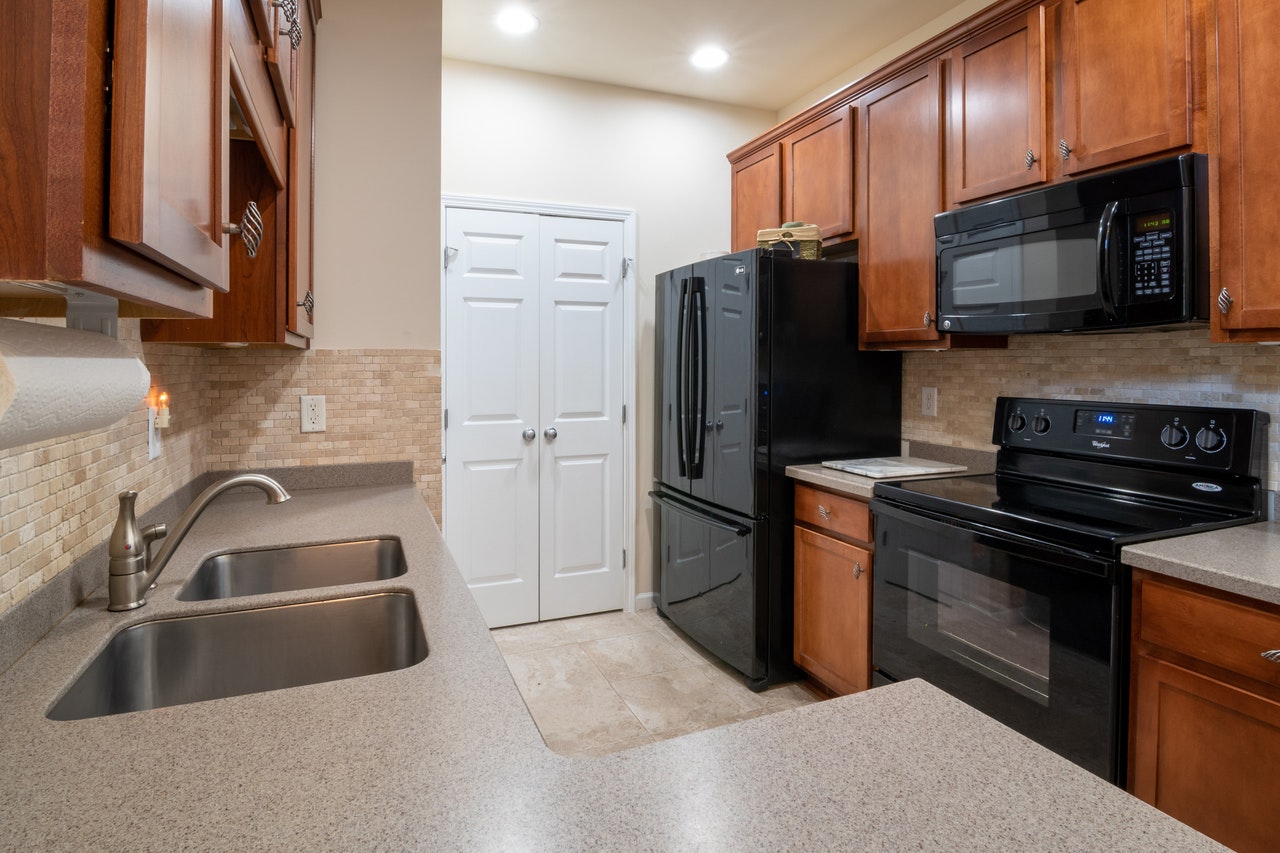} &
    \includegraphics[width=0.25\linewidth]{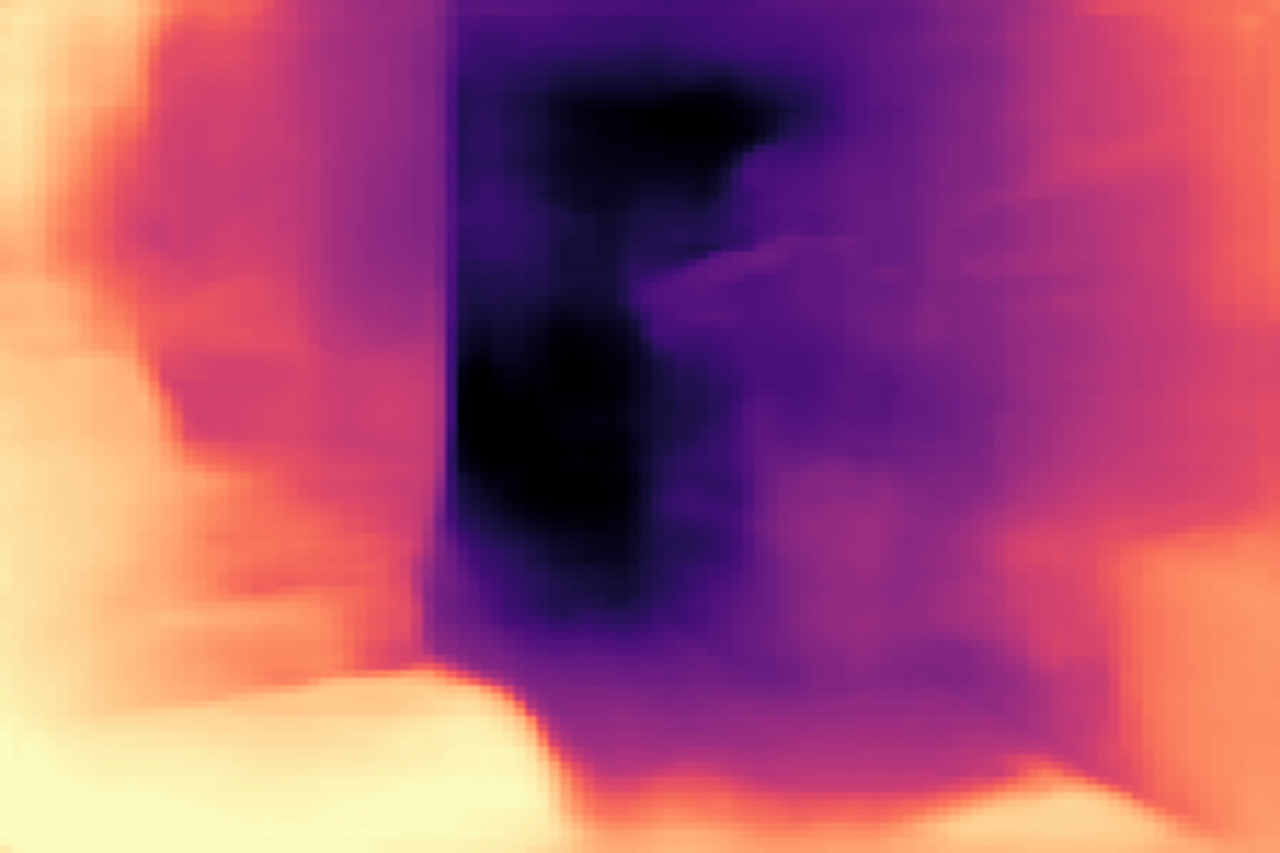} &
    \includegraphics[width=0.25\linewidth]{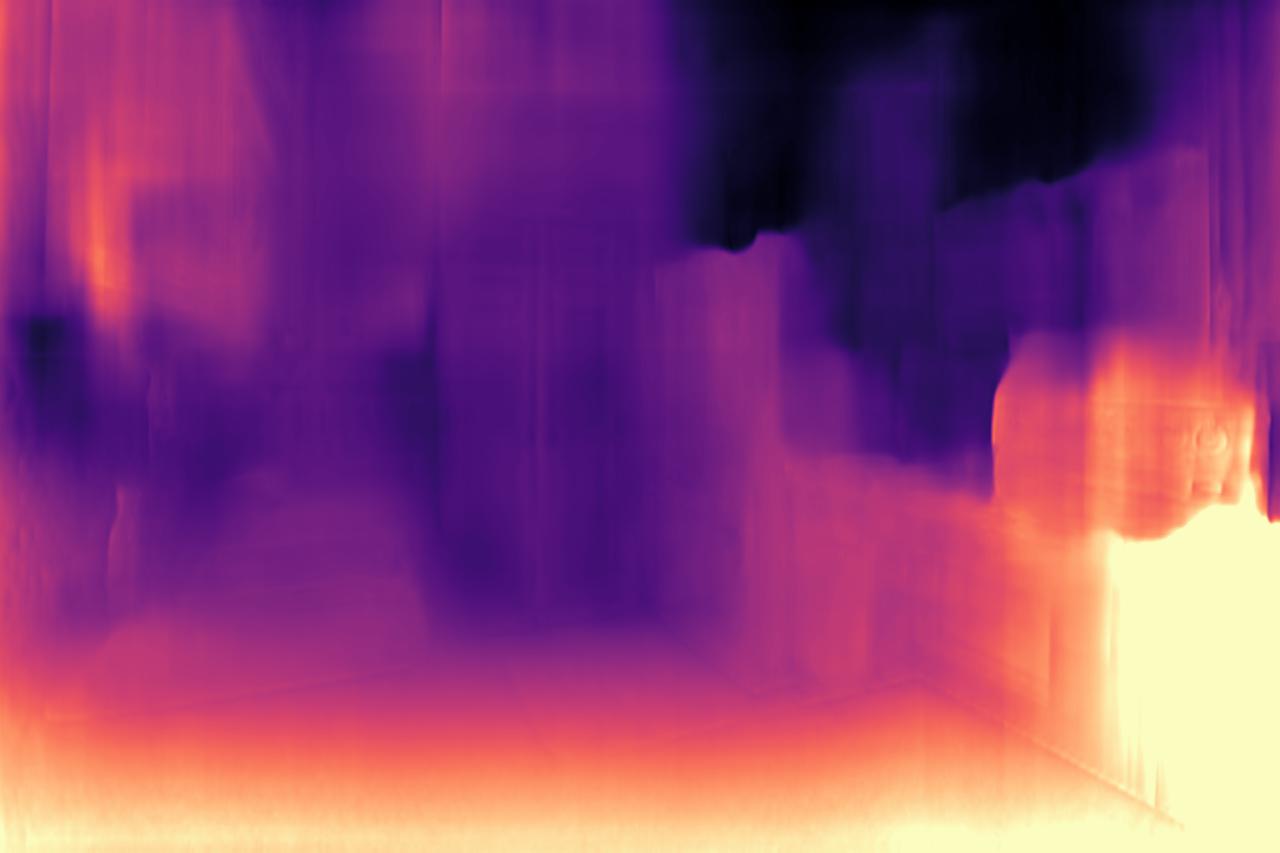} &
    \includegraphics[width=0.25\linewidth]{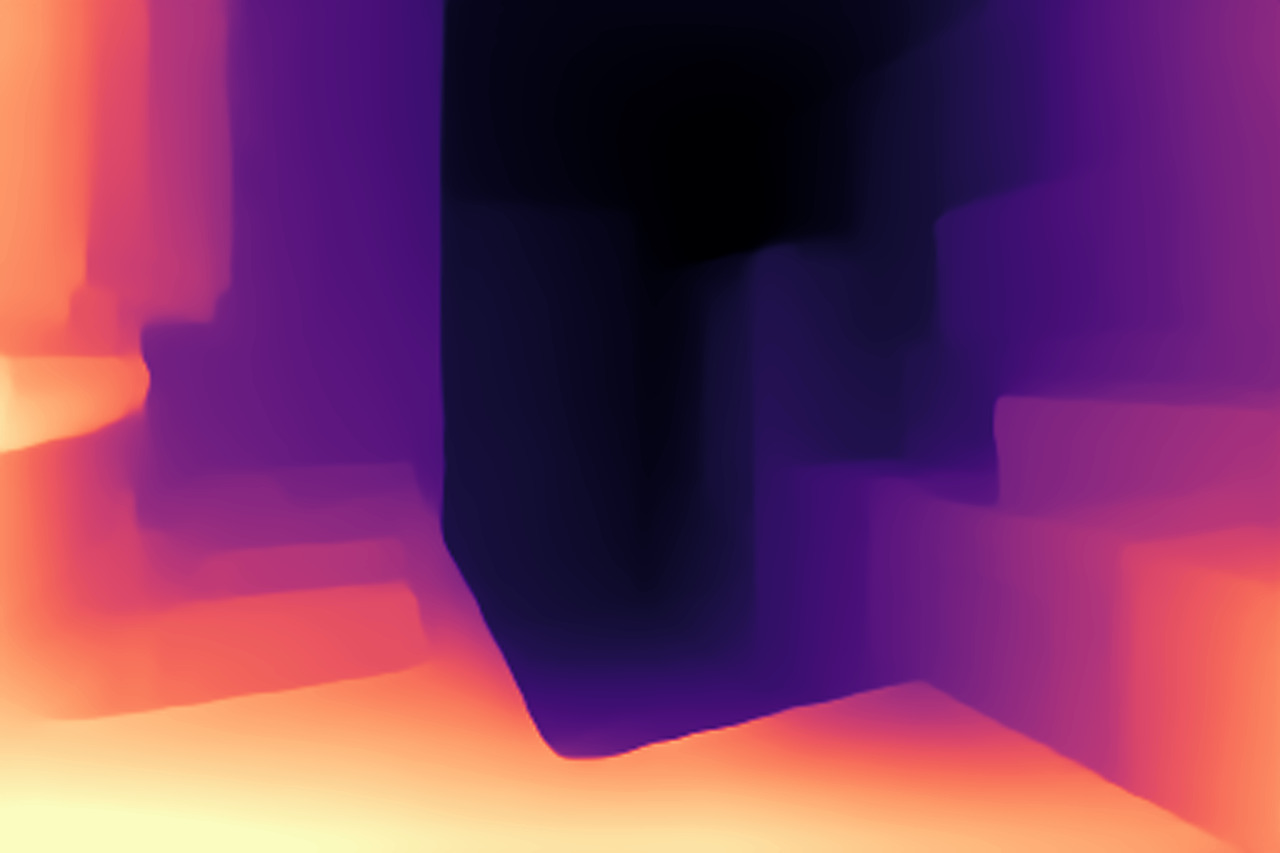}\\
    \includegraphics[width=0.25\linewidth]{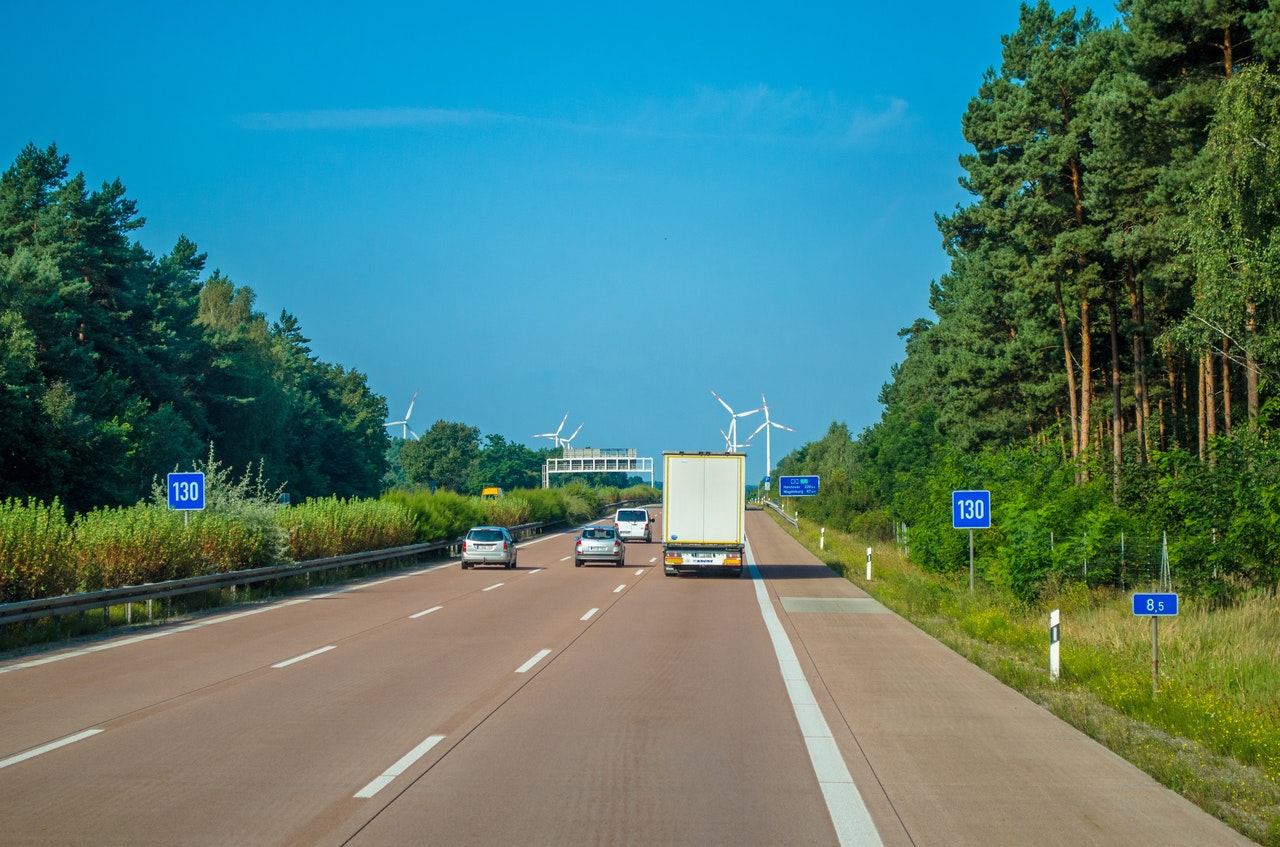} &
    \includegraphics[width=0.25\linewidth]{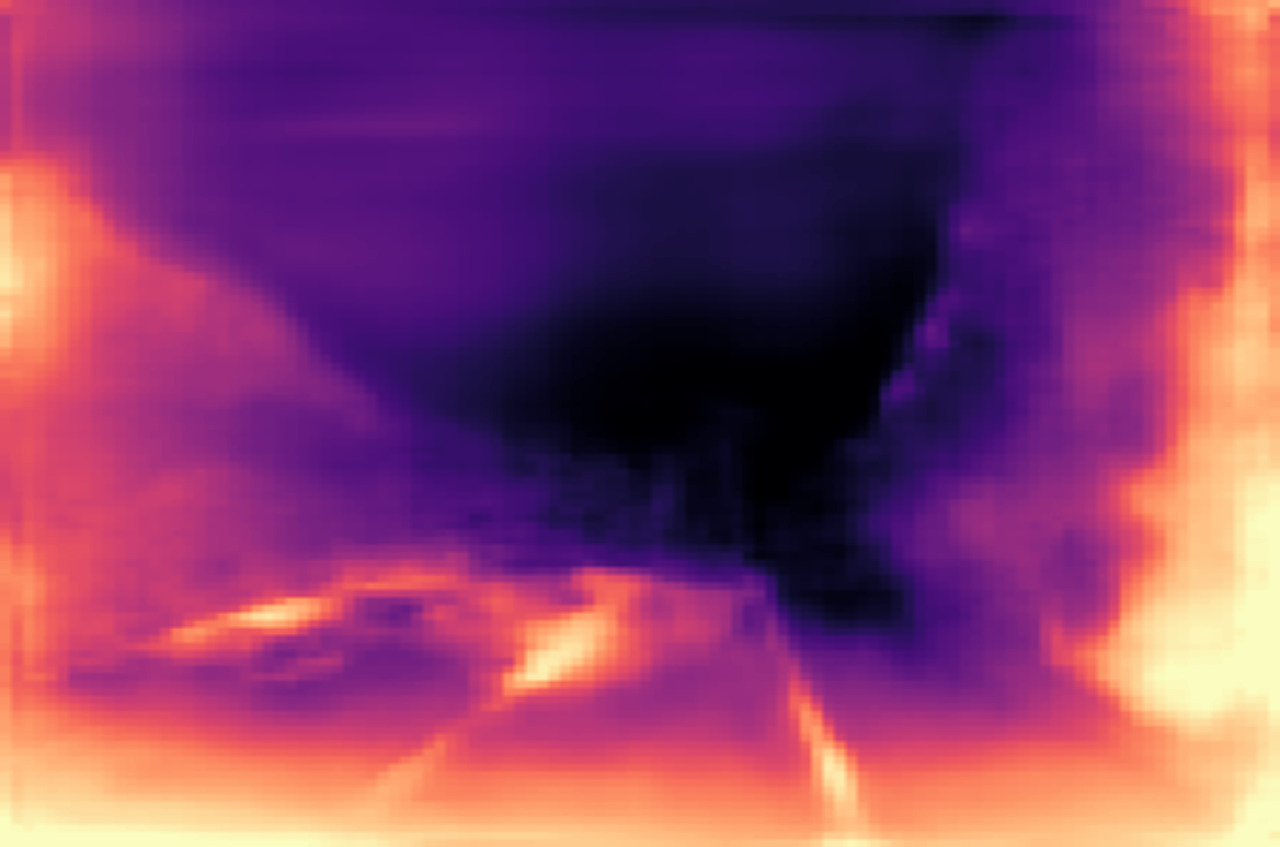} &
    \includegraphics[width=0.25\linewidth]{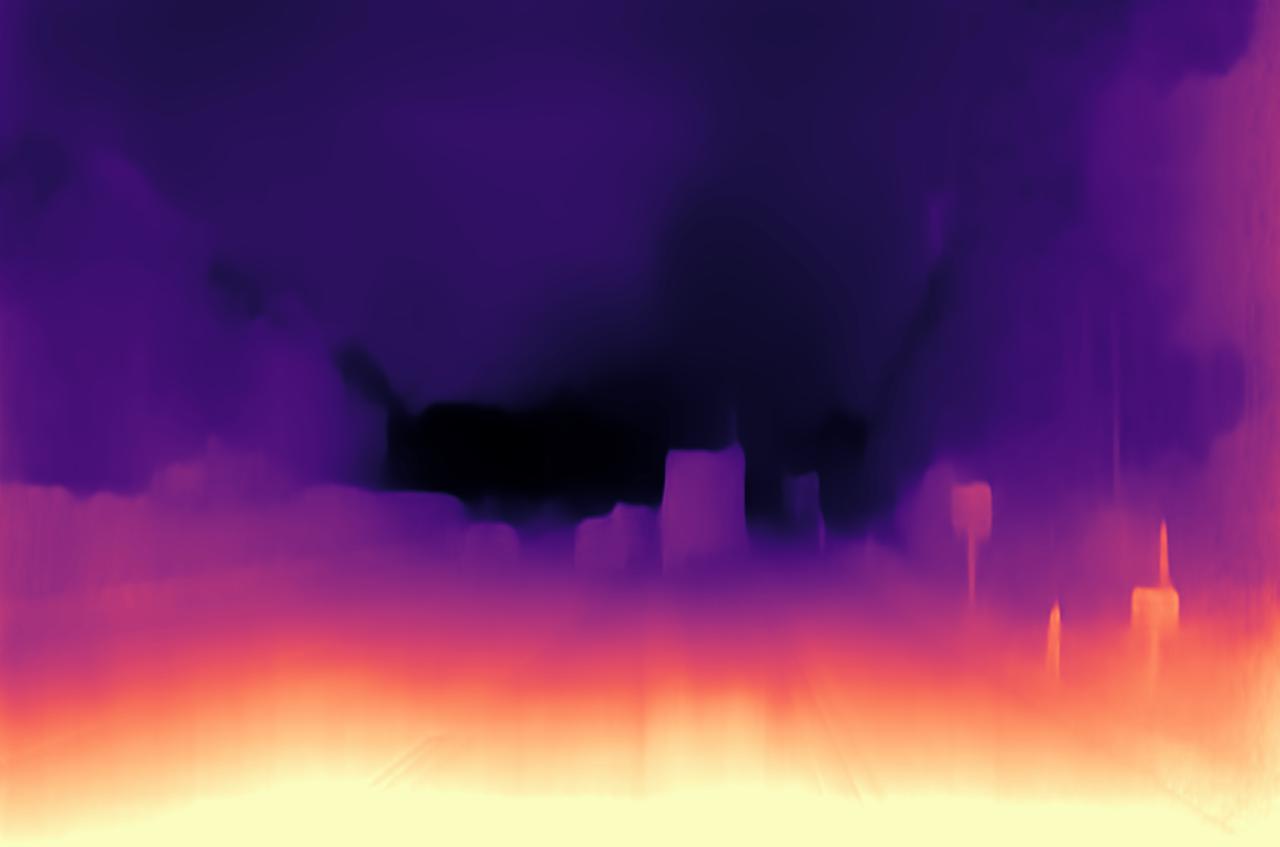} &
    \includegraphics[width=0.25\linewidth]{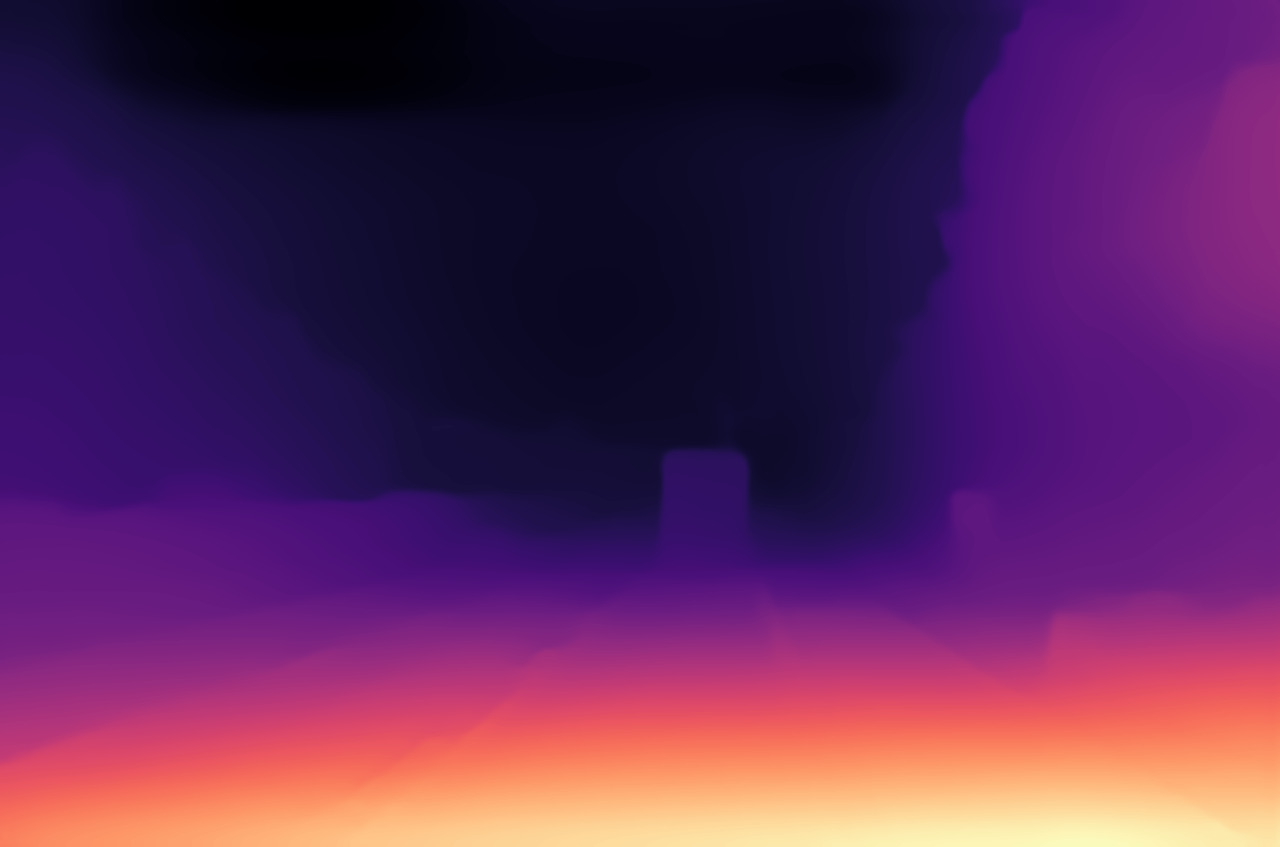}\\
    Reference & \fastdepth{} \cite{icra_2019_fastdepth} & Monodepth2 \cite{Godard_ICCV_2019} & MiDaS \cite{midas}
    \end{tabular}
    \caption{\textbf{Predictions in the wild.} We provide qualitative results for indoor and outdoor internet images. For each network, we used the checkpoints publicly available. It can be noticed how the networks trained on a single dataset, both in a supervised (\fastdepth{}) and self-supervised (Monodepth2), are not able to generalize well on a different setup. On the contrary, the network trained on various datasets (MiDaS) produces better results. Images from Pexels \url{https://www.pexels.com/}.}
    \label{fig:predictions_in_the_wild}
\end{figure*}

Considering that all the models have been trained with different configurations on different datasets, we re-train all the architectures exploiting the training framework of \cite{Godard_ICCV_2019} for a fair comparison. Specifically, we run 20 epochs of training for each model, decimating the learning rate after 15, on the Eigen train split of KITTI. We use Adam optimizer \cite{adam}, with an initial learning rate of $10^{-4}$, and minimize the highest three available scales for all the network except \fastdepth{}, which provides full-resolution (\ie{} $640\times192$) predictions only. Since the training framework expects a normalized inverse depth as the output of the network, we replace the last activation of each architecture (if present) with a sigmoid.

Table \ref{table:eigen} summarizes the experimental results of the models tested on the Eigen split of KITTI. The top four rows report the results, if available, provided in the original papers, while last three the accuracy of models re-trained within the framework described so far. This test allows for evaluating the potential of each architecture in fair conditions, regardless of the specific practices, advanced tricks or pre-training being deployed in the original works.
Not surprisingly, larger \resnet{} model performs better than the three lightweight models, showing non-negligible margins on each evaluation metric when trained in fair conditions. Among these latter, although their performance is comparable, \pyd{} results more effective with respect to \fastdepth{} and \dsnet{} on most metrics, such as RMSE and $\delta<1.25$.

Figure \ref{fig:eigen} shows some qualitative results, enabling us to compare depth maps estimated by the four networks considered in our evaluation on a single image from the Eigen test split.

\subsection{Evaluation in the wild} 

In the previous section, we have assessed the performance of the considered lightweight networks on a data distribution similar to the training one. Unfortunately, this circumstance is seldom found in most practical applications, and typically it is not known in advance where a network will be deployed. Therefore, how to achieve reliable depth maps in the wild? In Figure \ref{fig:predictions_in_the_wild} we report some qualitative results about original pre-trained networks on different scenarios. Notice that the first two networks have strong constraints about input size ($224\times 224$ for \cite{icra_2019_fastdepth}, $1024\times320$ for \cite{Godard_ICCV_2019}) that these networks internally apply, imposed by how these models have been trained in their original context. Although this limitation, \fastdepth{} (second column) can predict a meaningful result in an indoor environment (first row), not in outdoor (second row). It is not surprising since the network was trained on NYUv2, which is an indoor dataset. Monodepth2 \cite{Godard_ICCV_2019} suffers from the same problem, highlighting that this issue is not concerned with the network size (smaller the first, larger the second) or training approach (supervised the first, self-supervised the second), but it is rather related to the training data. Conversely, MiDaS by Ranftl \etal{} \cite{midas}, is effective in both situations. Such robustness comes from a  mixture of datasets, collecting about 2M frames covering many different scenarios, used to train a large ($\sim 105$M parameters) and very accurate monocular network.

\begin{table*}[!ht]
\centering
\setlength{\tabcolsep}{6pt}
\begin{tabular}{l|c|cccc|ccc}
\multicolumn{4}{c}{} & \multicolumn{2}{c}{\cellcolor{lower} Lower is better} & \multicolumn{2}{c}{\cellcolor{higher} Higher is better}\\
\hline
Network & Dataset & \cellcolor{lower}Abs Rel & \cellcolor{lower}Sq Rel & \cellcolor{lower}RMSE & \cellcolor{lower}log RMSE & \cellcolor{higher}$\delta<$1.25 &  \cellcolor{higher}$\delta<1.25^2$ & \cellcolor{higher}$\delta<1.25^3$  \\
\hline

Ranftl \etal{} \cite{midas} & TUM & \textbf{0.125}  &  \textbf{0.148}  &  \textbf{0.832}  &  \textbf{0.195}  & \textbf{0.857} & \textbf{0.944}  & \textbf{0.978}  \\
Li \etal{} \cite{li2019learning} &  TUM & 0.135  &   0.158  &   0.852  &   0.209  &   0.826  &   0.942  &   0.975 \\
\hline
\resnet{} & TUM &  \textcolor{red}{\textbf{0.147}}  &  \textcolor{red}{  \textbf{0.180}}  &   \textcolor{red}{  \textbf{0.916}}  &   \textcolor{red}{  \textbf{0.222}}  &   \textcolor{red}{  \textbf{0.811}}  &   \textcolor{red}{  \textbf{0.927} }  &   \textcolor{red}{ \textbf{0.967}} \\
\pyd{} & TUM & 0.166  &   0.210  &   0.978  &   0.244  &   0.767  &   0.921  &   0.955  \\ 
\dsnet{} & TUM &   0.168  &   0.215  &   0.994  &   0.248  &   0.762  &   0.917  &   0.951\\
\fastdepth{} & TUM  & 0.160  &   0.209  &   0.982  &   0.241  &   0.780  &   0.918  &   0.955 \\
\hline
\hline
Ranftl \etal{} \cite{midas}  & KITTI & \textbf{0.157}  &  \textbf{1.144}  &  \textbf{5.672}  &   \textbf{0.225}  &  \textbf{0.780}  &  \textbf{0.942}  &   \textbf{0.980} \\
Li \etal{} \cite{li2019learning} &  KITTI  & 0.227  &   2.081  &   7.841  &   0.325  &   0.621  &   0.854  &   0.939\\
\hline
\resnet{} & KITTI & 0.164  &   \textcolor{red}{  \textbf{1.194}}  &   6.000  &   \textcolor{red}{  \textbf{0.239}}  &   0.752  &   0.928  &   0.974\\
\pyd{} & KITTI  &   \textcolor{red}{  \textbf{0.162}}  &   1.272  &   6.138  &   \textcolor{red}{  \textbf{0.239}}  &   \textcolor{red}{  \textbf{0.760}}  &   0.927  &   0.974 \\
\dsnet{} & KITTI  & 0.164  &   1.203  &   \textcolor{red}{  \textbf{5.977}}  &   \textcolor{red}{  \textbf{0.239}}  &   0.754  &   \textcolor{red}{  \textbf{0.929}}  &   \textcolor{red}{  \textbf{0.975}} \\
\fastdepth{} & KITTI  &   0.168  &   1.227  &   6.017  &   0.241  &   0.741  &   0.927  &  \textcolor{red}{\textbf{0.975}} \\
\hline
\hline
Ranftl \etal{} \cite{midas} & NYU & \textbf{0.100}  & \textbf{0.061}  &   \textbf{0.407}  &  \textbf{ 0.132}  &  \textbf{ 0.905}  &  \textbf{ 0.984}  &   \textbf{0.997}  \\
Li \etal{} \cite{li2019learning} &  NYU & 0.149  &   0.116  &   0.560  &   0.189  &   0.782  &   0.958  &   0.992\\
\hline
\resnet{} & NYU  &  \textcolor{red}{\textbf{0.123}}  &  \textcolor{red}{\textbf{0.082}}  &  \textcolor{red}{\textbf{0.473}}  &   \textcolor{red}{\textbf{0.160}}  &  \textcolor{red}{\textbf{0.848}}  &  \textcolor{red}{\textbf{0.974}}  &  \textcolor{red}{\textbf{0.995}} \\
\pyd{} & NYU  &  0.130  &   0.091  &   0.493  &   0.168  &   0.827  &   0.969  &   0.994 \\
\dsnet{} & NYU  & 0.134  &   0.096  &   0.505  &   0.171  &   0.820  &   0.968  &   0.993 \\
\fastdepth{} & NYU  &  0.129  &   0.090  &   0.492  &   0.167  &   0.833  &   0.971  &   0.994 \\
\end{tabular}
\caption{\textbf{Evaluation of generalization capability.} The three groups from top to bottom report experimental results concerned, respectively, with (top) TUM dataset, (middle) KITTI Eigen and (bottom) NYUv2.}
\label{table:tum}
\end{table*}

\begin{figure*}[t]
    \centering
    \renewcommand{\tabcolsep}{1.0px}
    \begin{tabular}{cccccccc}
    \includegraphics[width=0.125\linewidth]{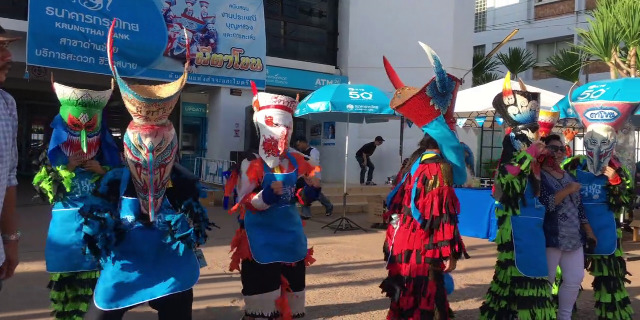} & 
    \includegraphics[width=0.125\linewidth]{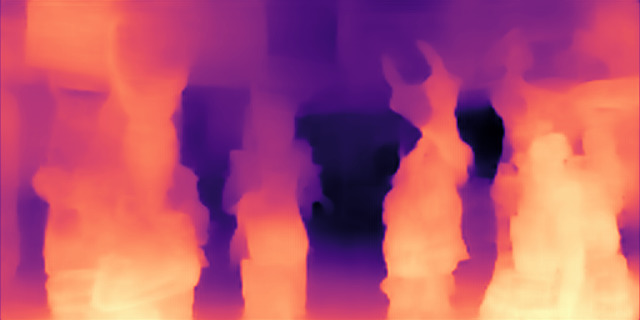} &
    \includegraphics[width=0.125\linewidth]{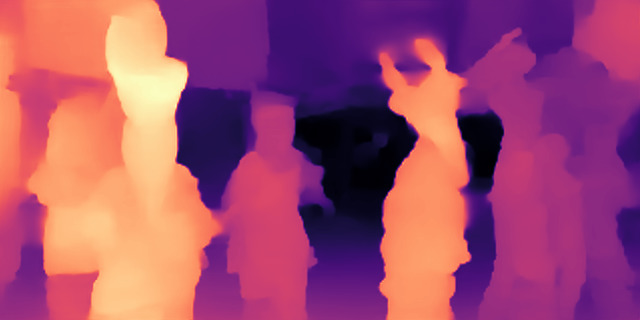} &
    \includegraphics[width=0.125\linewidth]{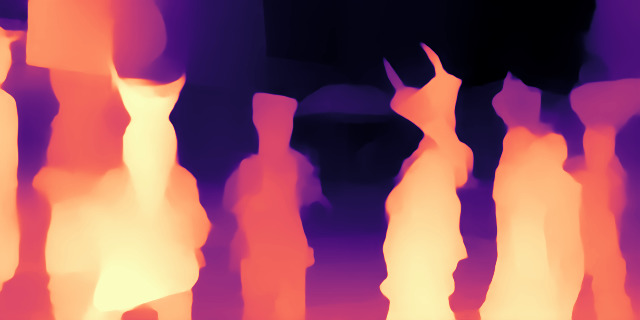} &
    \includegraphics[width=0.125\linewidth]{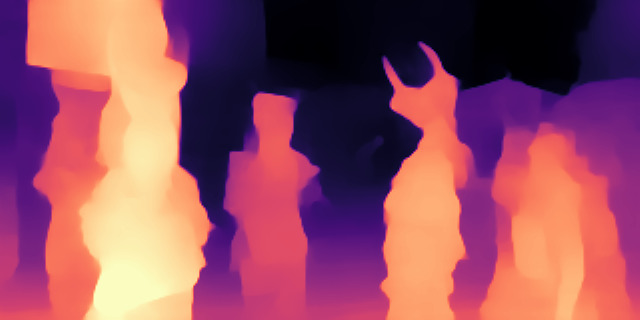} &
    \includegraphics[width=0.125\linewidth]{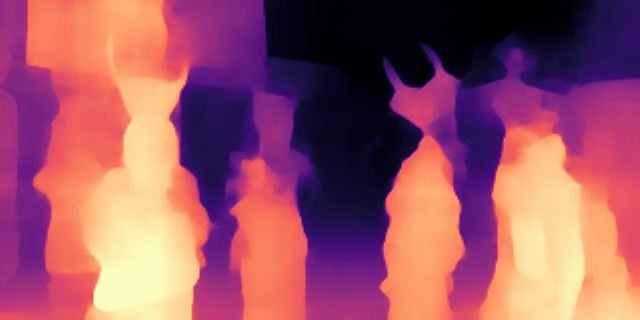} &
    \includegraphics[width=0.125\linewidth]{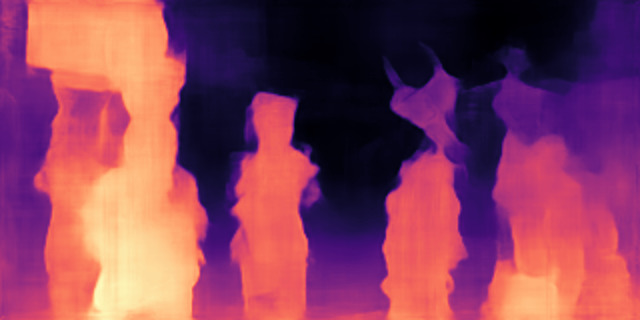} &
    \includegraphics[width=0.125\linewidth]{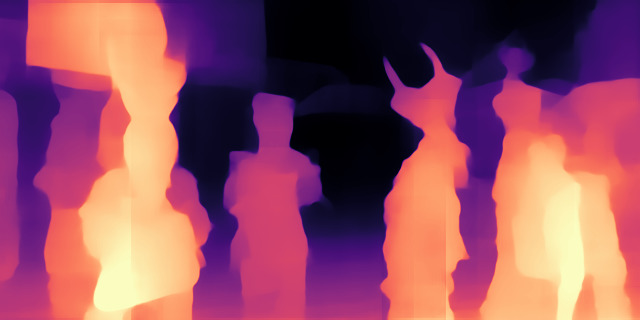} 
    \\
    \includegraphics[width=0.125\linewidth]{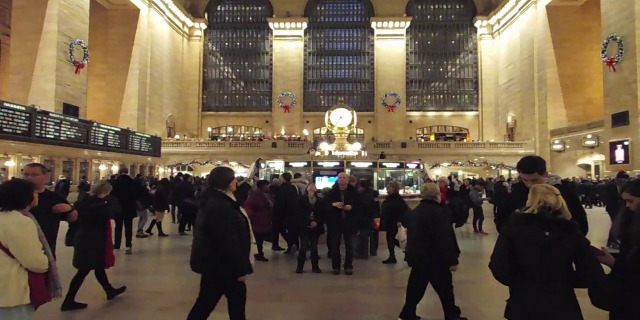} & 
    \includegraphics[width=0.125\linewidth]{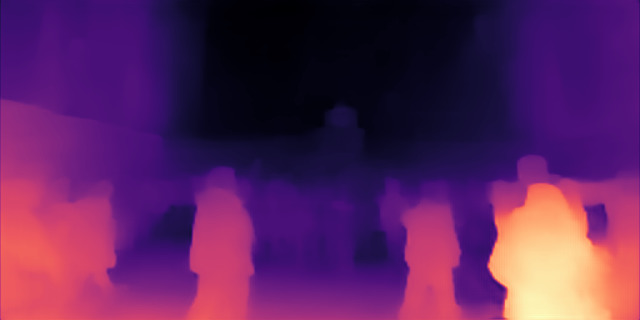} &
    \includegraphics[width=0.125\linewidth]{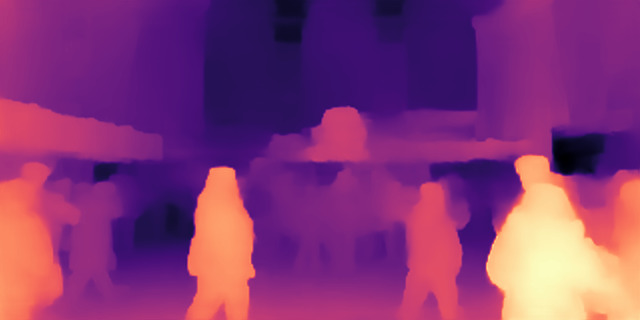} &

    \includegraphics[width=0.125\linewidth]{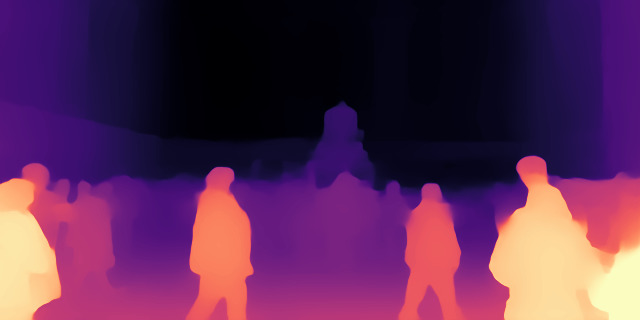} &
    \includegraphics[width=0.125\linewidth]{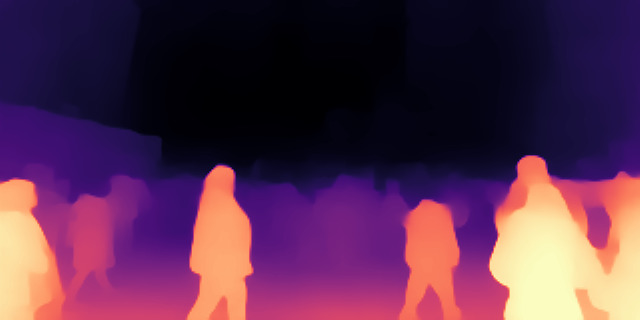} &
    \includegraphics[width=0.125\linewidth]{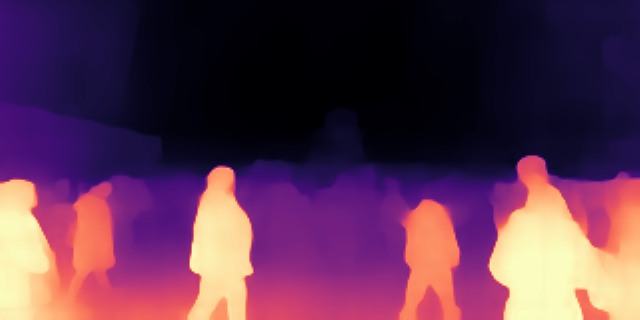} &
    \includegraphics[width=0.125\linewidth]{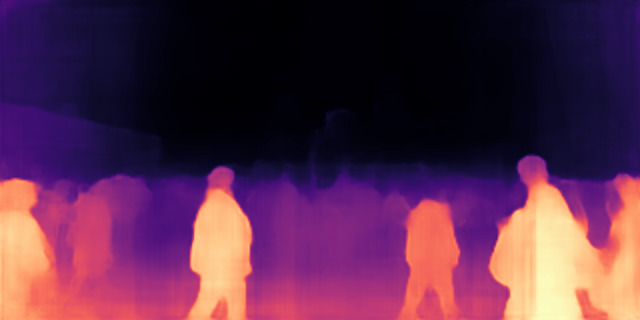} 
    &
    \includegraphics[width=0.125\linewidth]{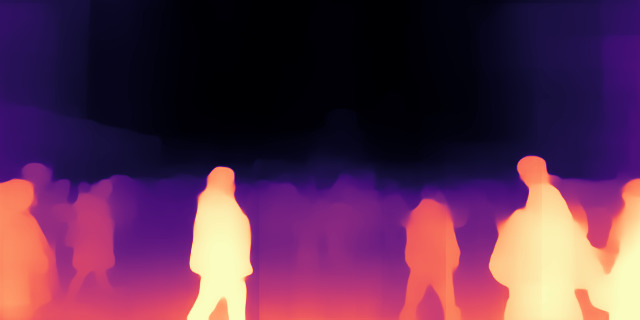} 
    \\
    \includegraphics[width=0.125\linewidth]{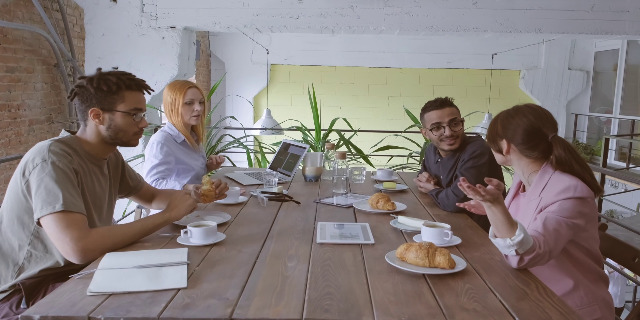} & 
    \includegraphics[width=0.125\linewidth]{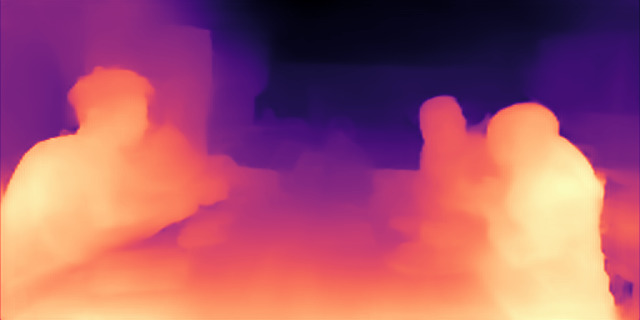} &
    \includegraphics[width=0.125\linewidth]{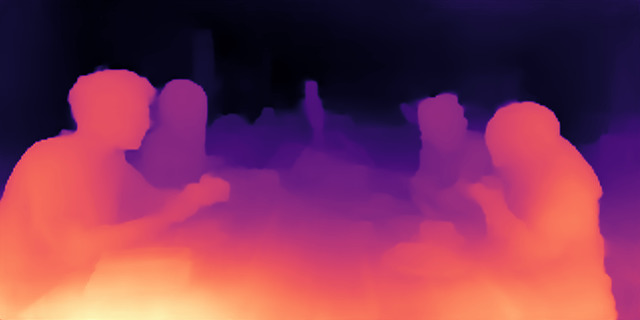} &
    \includegraphics[width=0.125\linewidth]{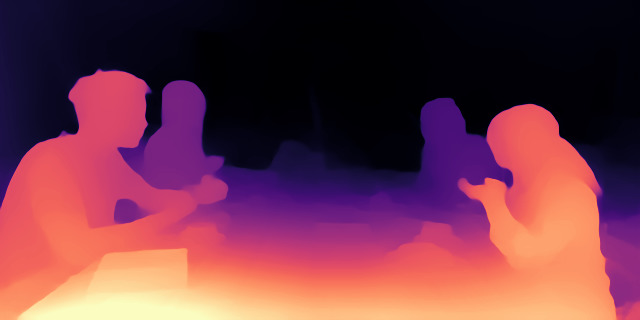} &
    \includegraphics[width=0.125\linewidth]{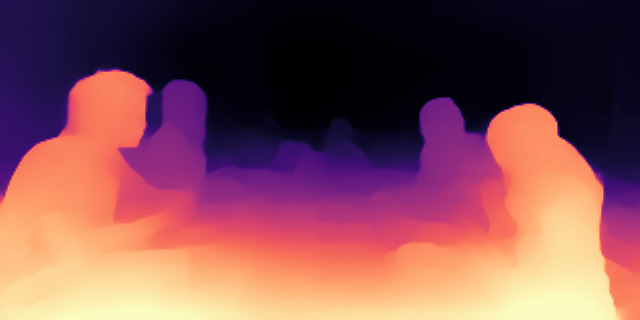} &
    \includegraphics[width=0.125\linewidth]{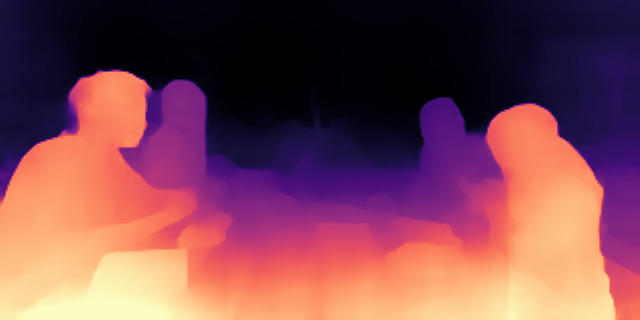}&
    \includegraphics[width=0.125\linewidth]{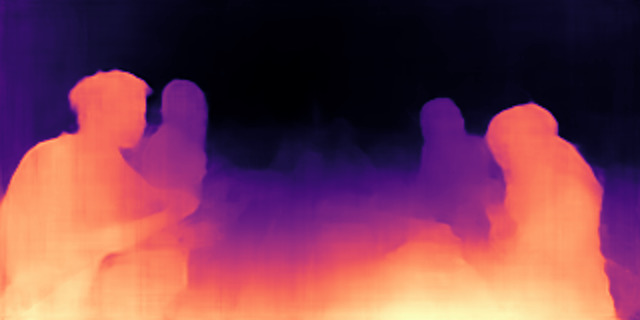}&
    \includegraphics[width=0.125\linewidth]{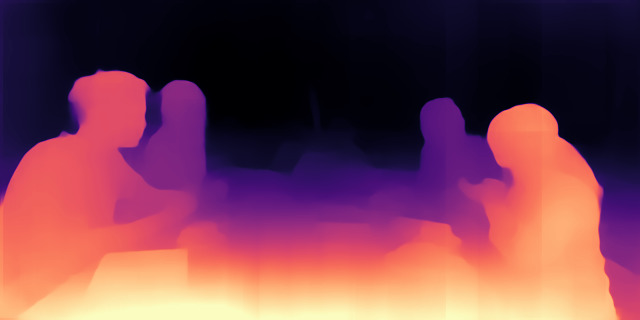}
    \\
    \includegraphics[width=0.125\linewidth]{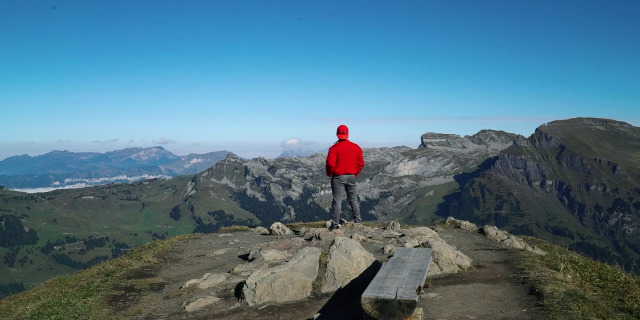} & 
    \includegraphics[width=0.125\linewidth]{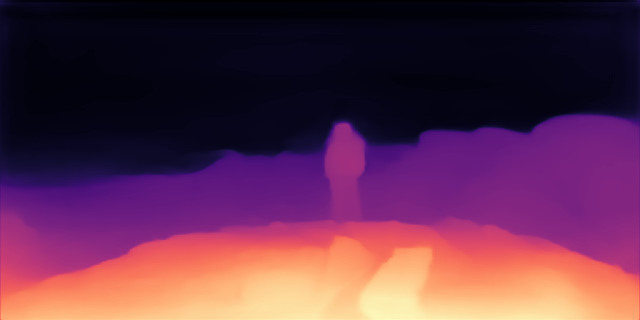} &
    \includegraphics[width=0.125\linewidth]{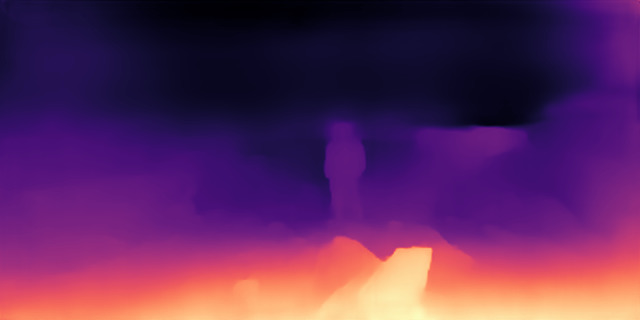} &
    \includegraphics[width=0.125\linewidth]{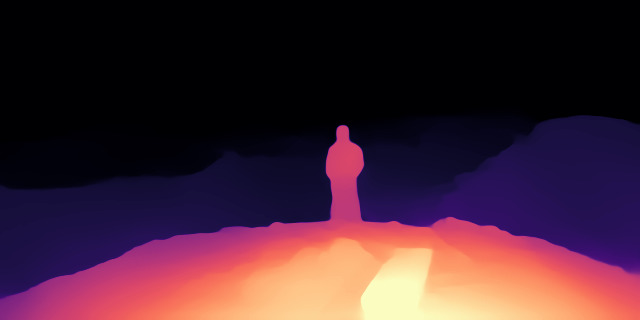} &
    \includegraphics[width=0.125\linewidth]{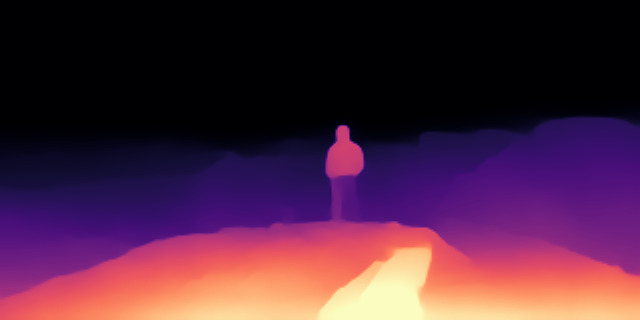} &
    \includegraphics[width=0.125\linewidth]{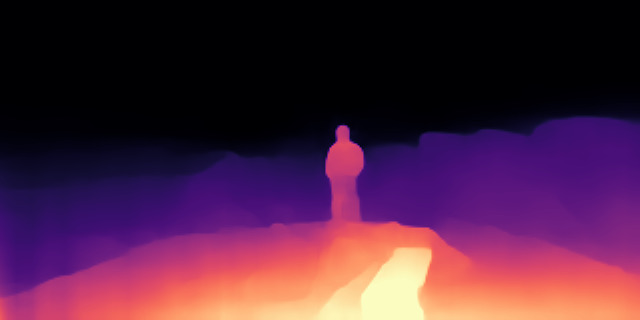}&
    \includegraphics[width=0.125\linewidth]{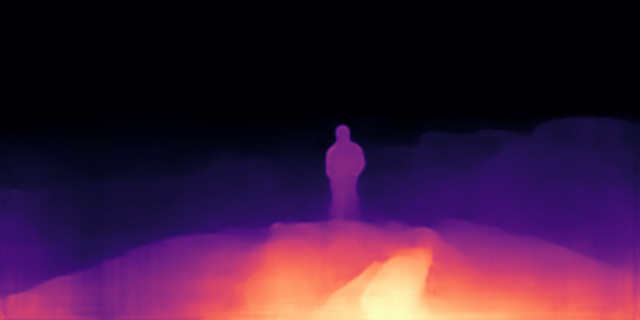}&
     \includegraphics[width=0.125\linewidth]{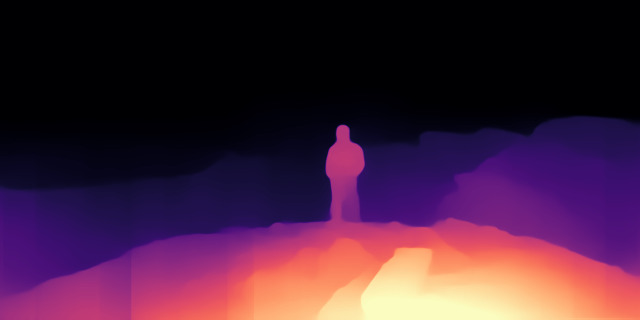}
    \\
    Reference & MegaDepth \cite{MegaDepthLi18} & Mannequin \cite{li2019learning} & MiDaS \cite{midas} & \pyd{} & \dsnet{} & \fastdepth{} & \resnet{} \\
    \end{tabular}
    \caption{\textbf{Qualitative results from internet photos}. From left to right, the reference image from Pexels website, depths from \cite{MegaDepthLi18}, \cite{midas} and our prediction, respectively for \pyd{} and \dsnet{}.}
    \label{fig:wild}
\end{figure*}

We leverage this latter model to \textit{distill} knowledge and train the lightweight models compatible with mobile devices. As mentioned before, this strategy allows use to \textit{use} MiDaS knowledge for faster training data generation compared to time-consuming pipelines used to train it, such as COLMAP \cite{schoenberger2016sfm,schoenberger2016mvs}. Moreover, it allows us to generate additional training samples and thus a much more scalable training set, potentially from any (single) image. Therefore, in order to train our network using the WILD dataset, we first generate proxy labels with MiDaS for each training image of this dataset. Then, obtained such proxy labels, we train the networks using the following loss function:

\begin{equation}\label{eq:loss_function}
 \mathcal{L}(D_x^s,D_{gt}) = \alpha_l \norm{(D_x^s - D_{gt})} + \alpha_s \mathcal{L}_g(D_x^s, D_{gt})
\end{equation}

where $\mathcal{L}_g$ is the gradient loss term defined in \cite{li2019learning}, $D_x^s$ the predictions of the network at scale $s$ (bilinearly upsampled to full resolution) and $D_{gt}$ is the proxy depth. The weight $\alpha_{s}$ depends on the scale $s$ and is halved at each lower scale. On the contrary, $\alpha_l$ is fixed and set to 1. Intuitively, the $L^1$ norm penalizes differences w.r.t proxies, while $\mathcal{L}_g$ helps to preserve sharp edges. We train the models for $40$ epochs, halving the learning rate after 20 and 30, with a batch size of 12 images, with an input size of $640\times320$. We set the initial value of $\alpha_s$ to $0.5$ for all networks except for \fastdepth{}, set to $0.01$. Additionally, for \resnet{} and \fastdepth{} feature upsampling through the nearest neighbour operator in the decoder phase have been replaced with bilinear interpolation. These changes were necessary to mitigate some checkboard artefacts found in depth estimations inferred by these networks following the training procedure outlined.

Table \ref{table:tum} collects quantitative results on three datasets, respectively TUM \cite{sturm12iros} (3D object reconstruction category), KITTI Eigen split \cite{Eigen_2014} and NYU \cite{Silberman:ECCV12}. On each dataset, we first show the results achieved by large and complex networks MiDaS \cite{midas} and the model by Li \etal{} \cite{li2019learning} (using the single frame version), both trained in the wild on a large variety of data. The table also reports results achieved by the four networks considered in our work trained on the WILD dataset exploiting knowledge distillation from MiDaS.
First and foremost, we highlight how MiDaS performs in general better than \cite{li2019learning}, emphasizing the reason to distil knowledge from it.

Considering lightweight compact models \pyd{}, \dsnet{} and \fastdepth{} we can notice that the margin between them and MiDaS is often non-negligible. Similar behaviour occurs for the significantly more complex network \resnet{} despite in general more accurate than other more compact networks, except on KITTI where it turns out less accurate when trained in the wild. However, considering the massive gap in terms of computational efficiency between compact networks and MiDaS analyzed later, that makes MiDaS not suited at all for real-time inference on the target devices the outcome reported in Table \ref{table:tum} is not so surprising.  
Looking more in details the outcome of lightweight networks, \pyd{} is the best model on KITTI when trained in the wild and also achieves the second-best accuracy on NYU, with minor drops on TUM. Finally, \dsnet{} and \fastdepth{} achieve average performance in general, never resulting in the best on any dataset.

Figure \ref{fig:wild} shows some qualitative examples of depth maps processed from internet pictures by MegaDepth \cite{MegaDepthLi18}, the model by Li \etal{} \cite{li2019learning}, MiDaS \cite{midas} and the fast networks trained through knowledge distillation in this work.

\begin{figure}[t]
    \centering
    \renewcommand{\tabcolsep}{1.0px}
    \begin{tabular}{ccc}
    \includegraphics[width=0.3\linewidth]{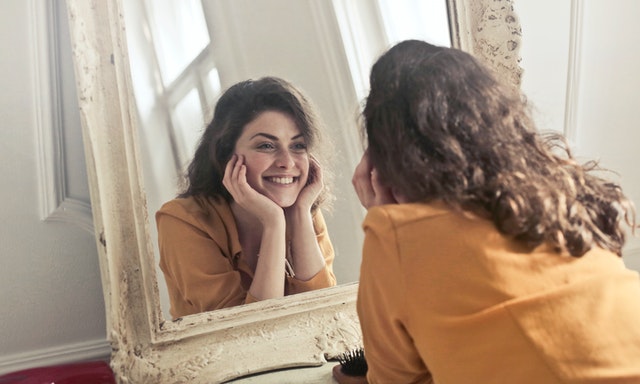} &
    \includegraphics[width=0.3\linewidth]{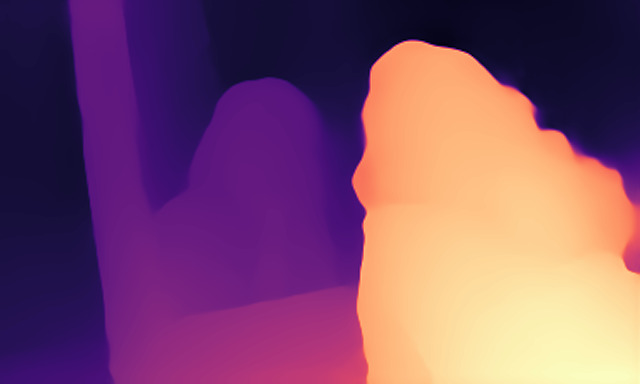} &
    \includegraphics[width=0.3\linewidth]{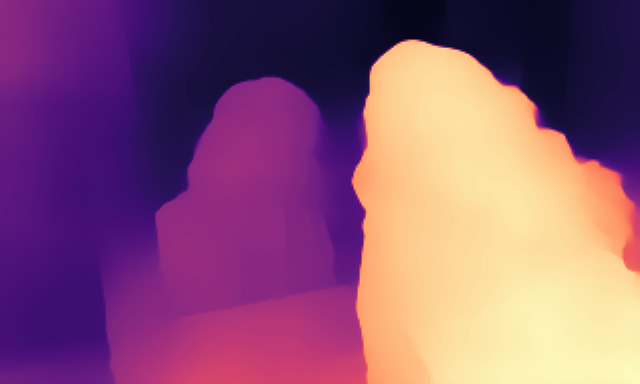} \\
    \includegraphics[width=0.3\linewidth]{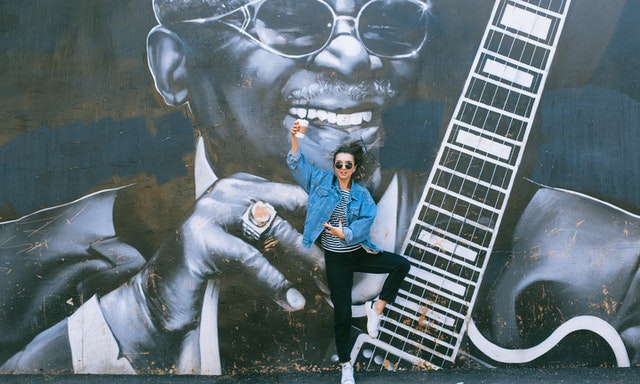} &
    \includegraphics[width=0.3\linewidth]{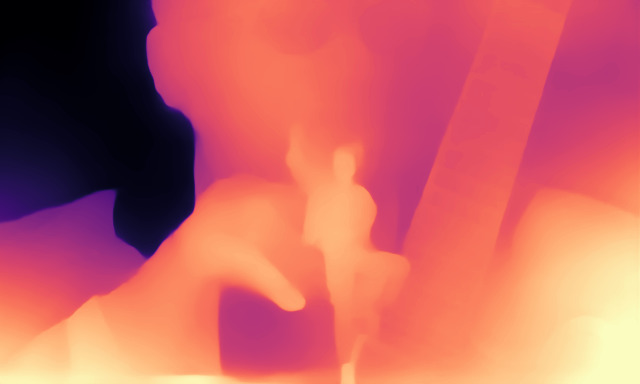} &
    \includegraphics[width=0.3\linewidth]{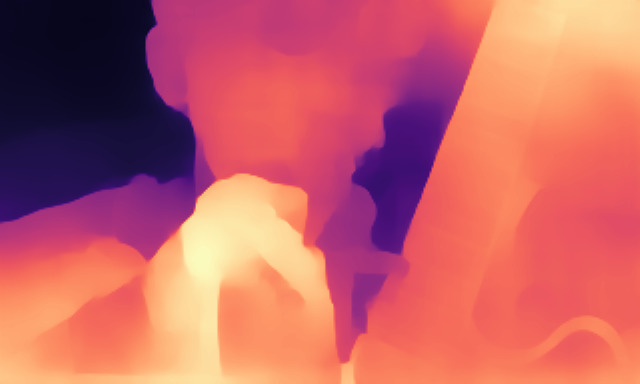} \\
    Reference & MiDaS & \pyd{} \\
    \end{tabular}

    \caption{\textbf{Failure cases.} Example of failure cases of single image depth estimation. From left to right: input image, depth predicted by the teacher and by the student.}
    \label{fig:failures}
\end{figure}

Finally, in Figure \ref{fig:failures} we report some example of failure cases of MiDaS (in the middle column) inherited by student networks. Since both networks fail, the problem is not attributable to their different architecture. Observing the figure, we can notice that such behavior occurs in very ambiguous scenes such as when dealing with mirrors or flat surfaces with content aimed at inducing optical illusions in the observers.

\begin{table}[t]
    \centering
    \begin{tabular}{cc|cc|}
        \hline
        Network & MAC (G) & FPS \\
        \hline
        MiDaS  & 172.4 & 0.20 \\
        \hline
        \resnet{} & 16.01 & 9.94\\
        \dsnet{} & 9.48 & 11.05 \\
        \pyd{} & 9.25 & 58.86 \\
        \fastdepth{} & 3.61 & 50.31\\
        
    \end{tabular}
    \caption{\textbf{Performance on smartphones}. We measure both the number of  \textit{multiply–accumulate operation} (MAC) and the FPS of monocular networks on an iPhone XS, using an input size of $640\times384$, averaged on 50 inferences.}
    \label{tab:smartphone_performances}
\end{table}

\subsection{Performance analysis on mobile devices}

After training the considered architectures on the WILD dataset, the stored weights can be converted into mobile-friendly models using tools provided by deep learning frameworks. Moreover, as previously specified, in our experiments, we perform only model conversion avoiding weights quantization not to alter the accuracy of the original network. 

Table \ref{tab:smartphone_performances} collects stats about the considered networks. Specifically, we report the number of \textit{multiply-accumulate} operations (MAC), computed with TensorFlow utilities, and the frame rate (FPS) measured when deploying the converted models on an Apple iPhone XS. Measurements are gathered processing $640\times384$ images and averaging over 50 consecutive inferences.
On top, we report the performance achieved by MiDaS, showing that it requires about 5 seconds on a smartphone to process a single depth map, performing about 170 billion operations. This evidence highlights how, despite being much more accurate, as shown before, this vast network is not suited at all for real-time processing on mobile devices.
Moving on more compact models, we can notice how \resnet{} reaches nearly 10 FPS performing one order of magnitude fewer operations. \dsnet{} and \pyd{} both perform about 9 billion operations, but the latter allows for much faster inference, close to 60 FPS and about 6 times faster than previous models.
Since the number of operations is almost the same for \dsnet{} and \pyd{}, we reconduct this performance discrepancy to low-level optimization of some specific modules.
Finally, \fastdepth{} performs 3 times fewer operations, yet runs slightly slower than \pyd{} when deployed with the same degree of optimization of the other networks on the iPhone XS.

Summarizing the performance analysis reported in this section and the previous accuracy assessment concerning the deployment of single image depth estimation in the wild, our experiments highlight \pyd{} as the best trade-off between accuracy and speed when targeting embedded devices. 

A video showing the deployment of \pyd{} with an iPhone XS framing an urban environment is available at \url{youtube.com/watch?v=LRfGablYZNw}. \\
At the following link is also available a \pyd{} web demo with client-side inference carried out by TensorFlow JS: \url{filippoaleotti.github.io/demo_live}.

\begin{figure}[t]
     \renewcommand{\tabcolsep}{1.2px}
    \centering
    \begin{tabular}{ccc}
    \includegraphics[width=0.33\linewidth]{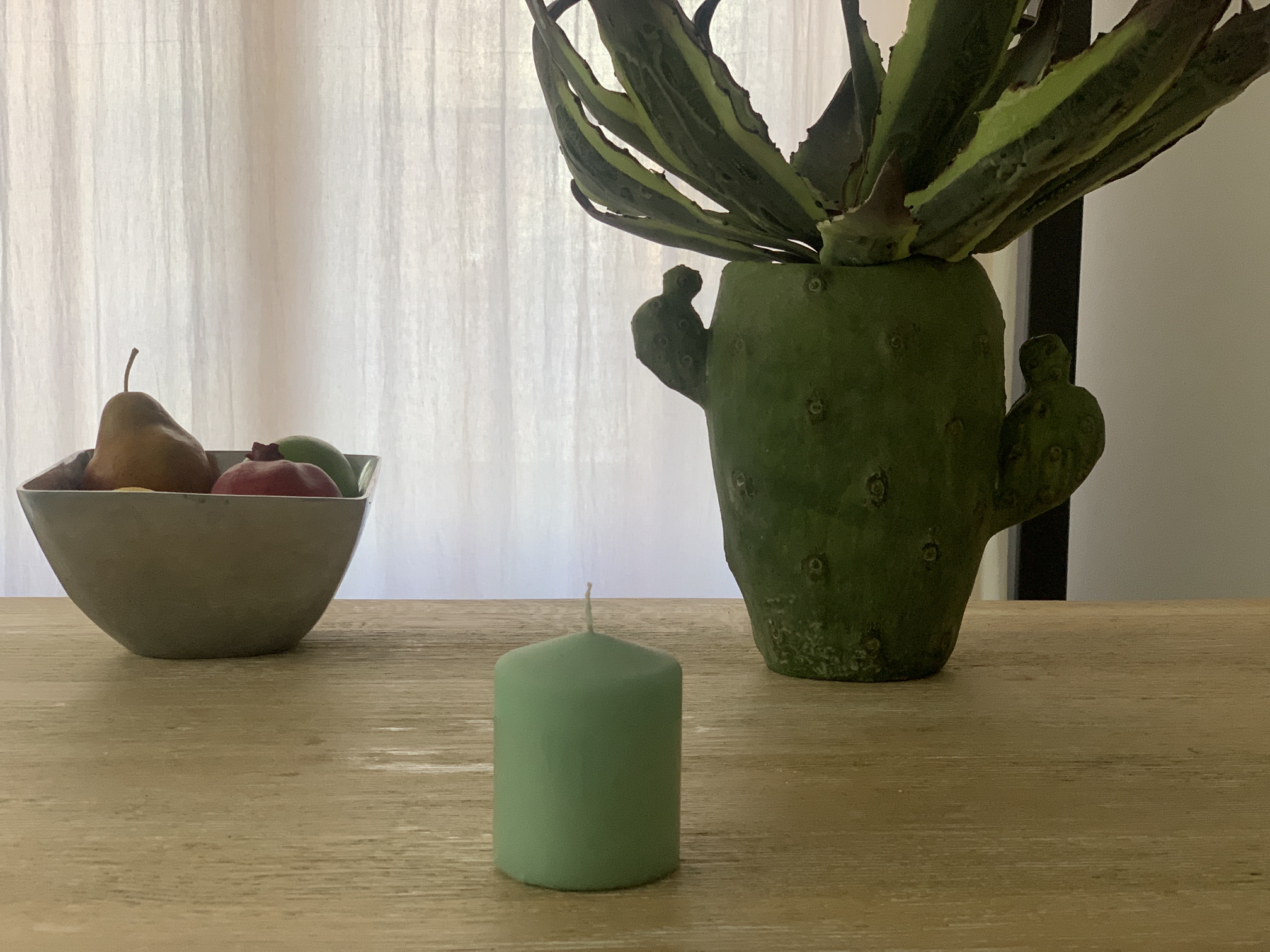} &
    \includegraphics[width=0.33\linewidth]{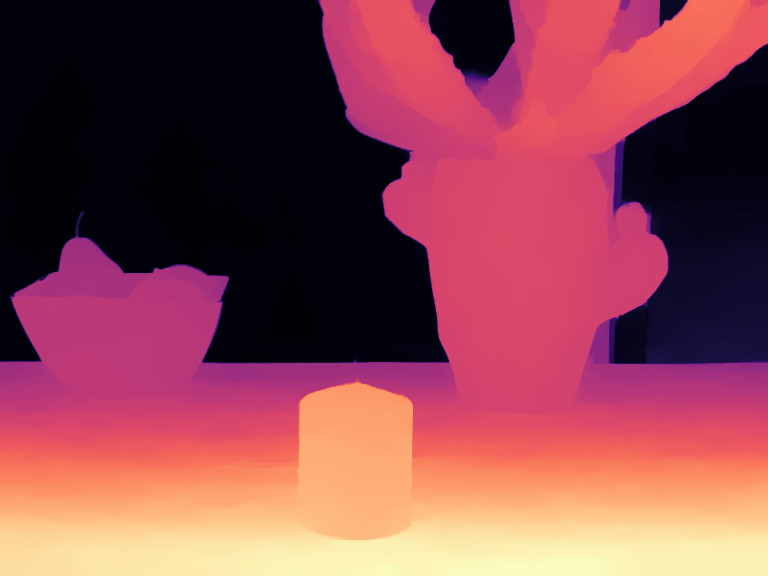} &
    \includegraphics[width=0.33\linewidth]{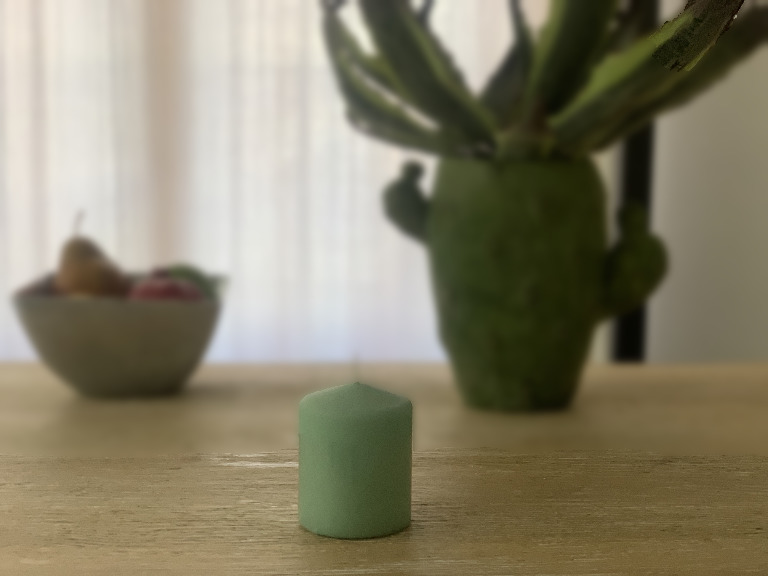} \\
    Reference & iOS stereo depth & Bokeh (stereo) \\
    \includegraphics[width=0.33\linewidth]{images/bokeh/reference.jpeg} &
    \includegraphics[width=0.33\linewidth]{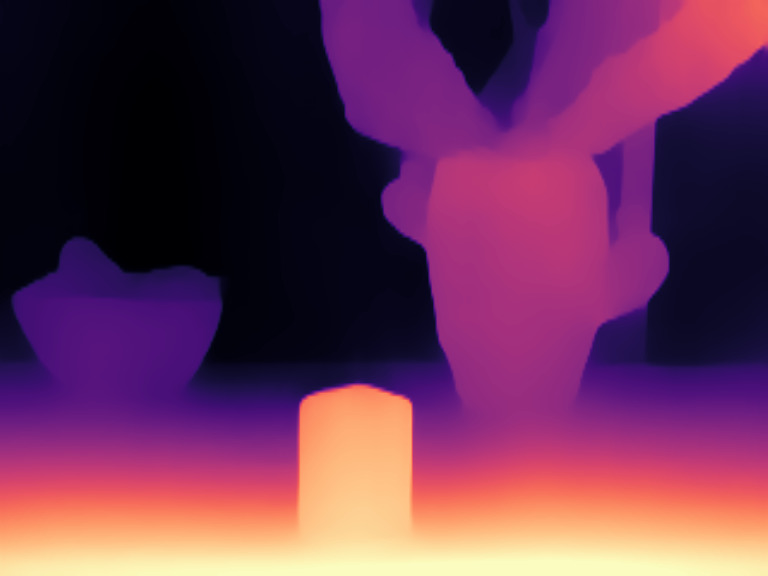} &
    \includegraphics[width=0.33\linewidth]{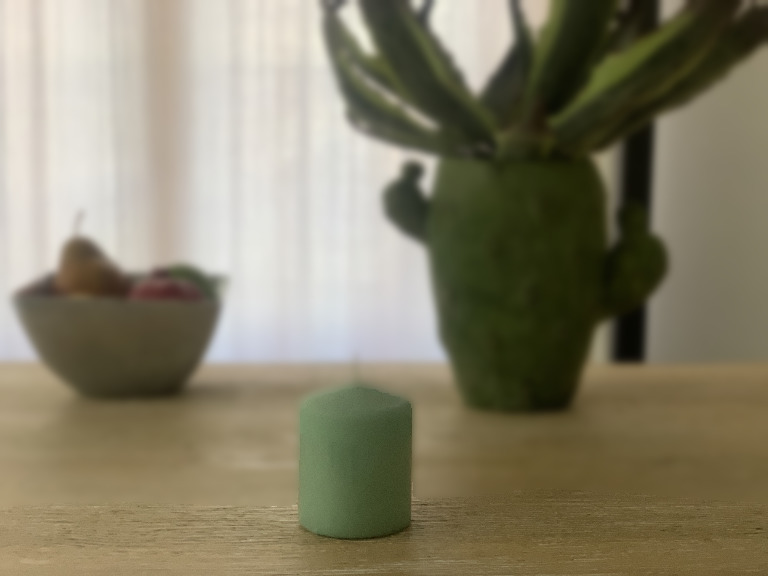} \\
    Reference & Pydnet depth & Bokeh (monocular) \\
    \includegraphics[width=0.33\linewidth]{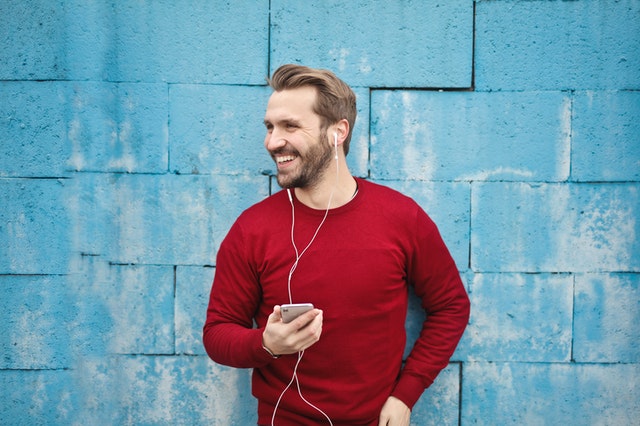} & 
    \includegraphics[width=0.33\linewidth]{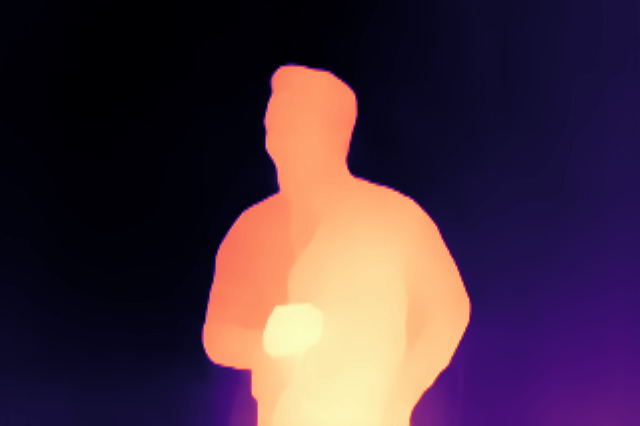} &
    \includegraphics[width=0.33\linewidth]{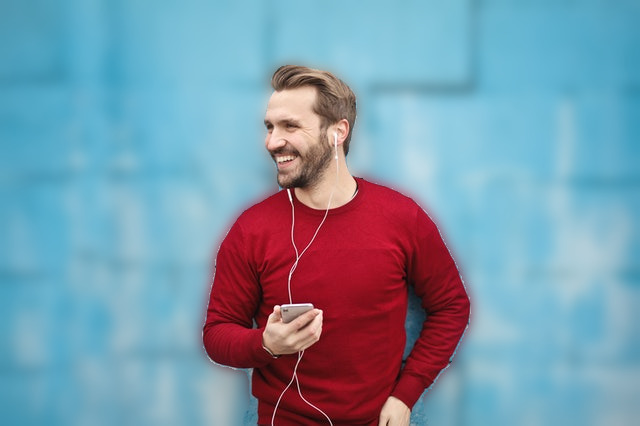} \\
    Reference & Pydnet depth & Bokeh (monocular)
    \end{tabular}
    \caption{\textbf{Bokeh effect.} Given the reference image acquired with an iPhone XS, we smooth farther pixels in the image using depth values provided by the native stereo method (first row) and \pyd{} monocular network (second row), obtaining similar results.
    Moreover, we can apply the same effect with \pyd{} using images from the web (third row).}
    \label{fig:bokeh}
\end{figure}

\section{Applications of single image depth estimation}

Once exhaustively assessed the performance of the considered lightweight networks, we present two well-known applications that can significantly take advantage of real-time and accurate single image depth estimation. For these experiments, we use the \pyd{} model trained on the WILD dataset, as described in previous sections.

\textbf{Bokeh effect.} The first application consists of a \textit{bokeh filter}, aimed at blurring an image according to the distance from the camera. More precisely, in our implementation, given a threshold $\tau$, all the pixels with a relative inverse depth larger than $\tau$ are blurred by a 25$\times$25 Gaussian kernel.

For our experiments, we captured a stereo pair using the rear cameras of an iPhone XS, and then using its API, we inferred a depth map to obtain a baseline. We also fed the \pyd{} network the single reference image of the stereo pair. Figure \ref{fig:bokeh} depicts the depth maps inferred by the stereo and monocular approach and the outcome of the bokeh filter. From the figure, we can notice that even if the depth map inferred by the monocular system is not in scale as the stereo one, it preserves pretty well details and also allows to retrieve the relative distance of objects. This latter feature, combined with the need for a single image is highly desirable is many consumer applications like the one described.
Additionally -- since the distance between the two imaging sensors of a stereo setup of a mobile phone is short-- the parallax effect enabling to infer depth with stereo vanishes close to the camera. On the contrary, a monocular system is agnostic to this problem. For this experiment, we set $\tau$ equals to $0.9$ and $0.7$ for stereo and monocular depth, respectively. 

Finally, another advantage consists of enabling the bokeh effect even when a stereo pair is not available, for instance, dealing with images sampled from the web (third row in Figure \ref{fig:bokeh}). From it, we can notice how processing a single input image enables a charming effect. Notice that bokeh effect with stereo is not applicable in this case since the stereo pair is not available as frequently occurs in practice.

\begin{figure}[t]
    \centering
    \renewcommand{\tabcolsep}{1.0px}
    \begin{tabular}{ccc}
    \includegraphics[width=0.33\linewidth]{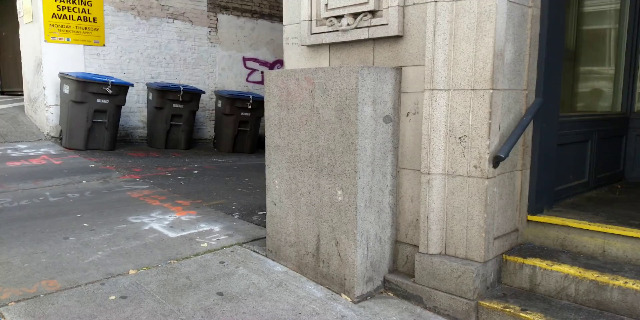} &
    \includegraphics[width=0.33\linewidth]{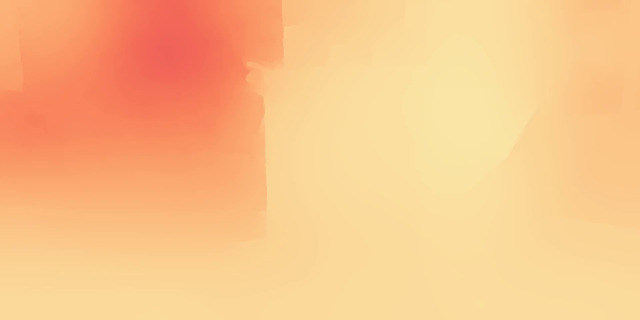} &
    \includegraphics[width=0.33\linewidth]{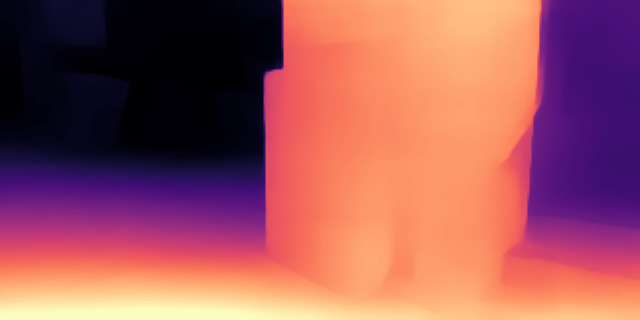} \\
    \includegraphics[width=0.33\linewidth]{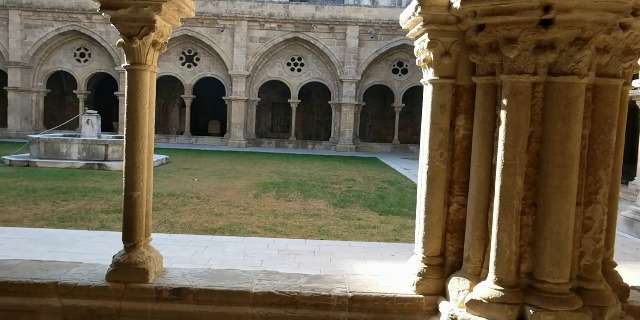} &
    \includegraphics[width=0.33\linewidth]{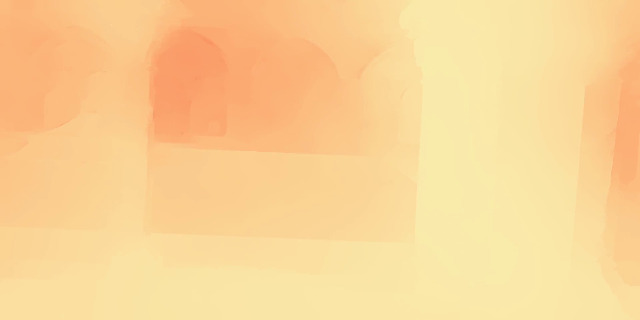} &
    \includegraphics[width=0.33\linewidth]{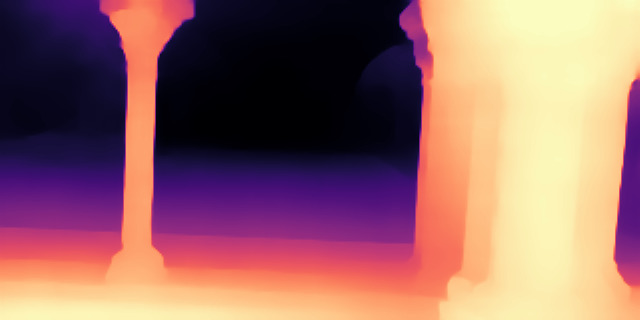} \\
    \includegraphics[width=0.33\linewidth]{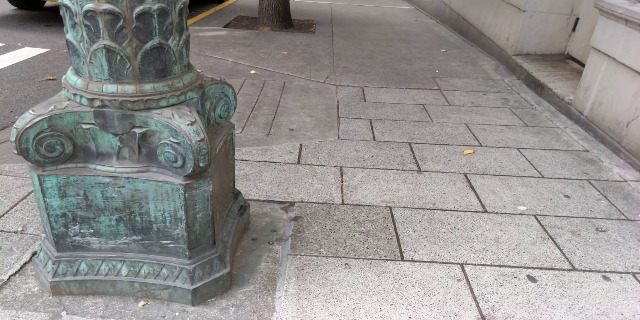} &
    \includegraphics[width=0.33\linewidth]{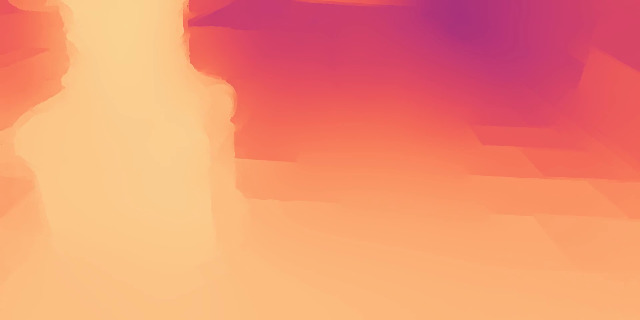} &
    \includegraphics[width=0.33\linewidth]{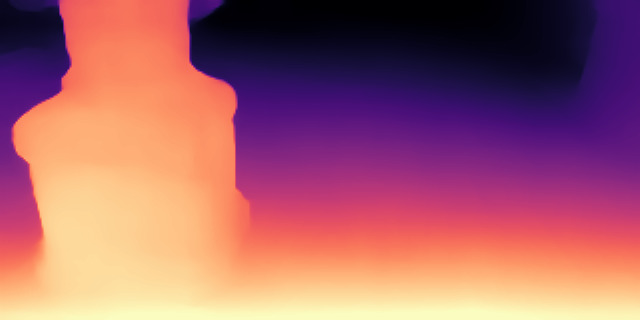} \\
    \includegraphics[width=0.33\linewidth]{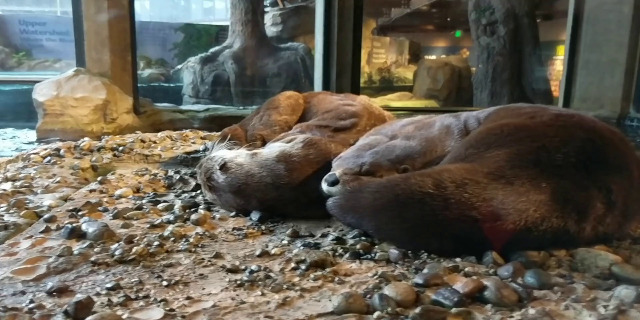} &
    \includegraphics[width=0.33\linewidth]{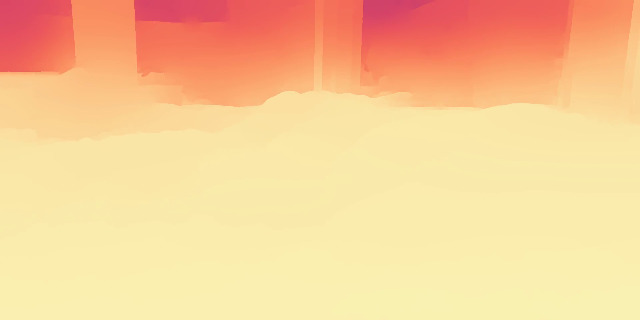} &
    \includegraphics[width=0.33\linewidth]{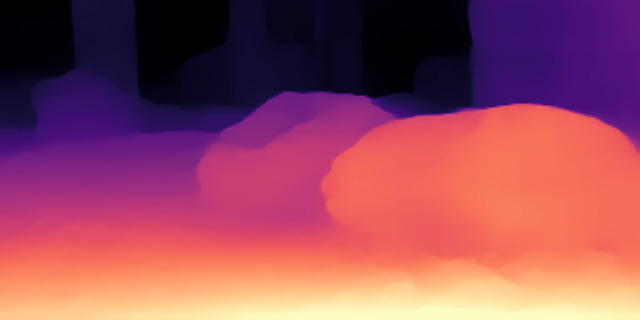} \\
   \\
    \end{tabular}
    \caption{\textbf{Qualitative comparison with other occlusion-aware AR methods.} From left to right, the input image, the depth from \cite{Occlusion2018} and \pyd{} predictions. 
    }
    \label{fig:ar_comparison}
\end{figure}

\textbf{Augmented reality with depth-aware occlusion handling.} Modern augmented reality (AR) frameworks for smartphones allows robust and consistent integration of virtual objects on flat areas leveraging camera tracking. However, they miserably fail when the scene contains occluding objects protruding from the flat surfaces. Therefore, in AR scenarios, dense depth estimation is paramount to handle properly physic interactions with the real world, such as occlusions. Unfortunately, most methods rely only on sparse depth measurements for a few points in the scene appropriately scaled exploiting the sensor suite of modern smartphones comprising accelerometers, gyroscope, etcetera. Although some authors proposed to densify such sparse measurements, it is worth observing that dynamic objects in the sensed scene may yield incorrect, sparse estimation and thus these methods need to filter out moving points \cite{Occlusion2018}.  
We argue that single image depth estimation may enable full perception of the scene suited for many real-world use cases potentially avoiding at all the issues outlined so far. The only remaining issue, concerned with the unknown scale factor intrinsic in a monocular system can be robustly addressed leveraging, as described next, one of the multiple sparse depth measurements in scale among those made available by standard AR frameworks. 

Purposely, we developed a mobile application capable of handling in real-time object occlusions by combining AR frameworks, such as ARCore or ARKit, and a robust and lightweight monocular depth estimation. network
To achieve this goal, we first exploit the AR framework to retrieve low-level information, such as the pose of the camera and the position of anchors in the sensed environment. Then, we retrieve depth for anchor points and, by comparing such measurements with monocular depth predictions, we can prevent to render occluded regions. At each frame, the scale factor issue is tackled within a robust RANSAC-based framework fed with the sparse and potentially noisy depth measurements provided by the AR framework and the dense depth map estimated by the monocular network.

\begin{figure}[t]
    \centering
    \renewcommand{\tabcolsep}{1.0px}
    \begin{tabular}{cc}
    \includegraphics[width=0.45\linewidth]{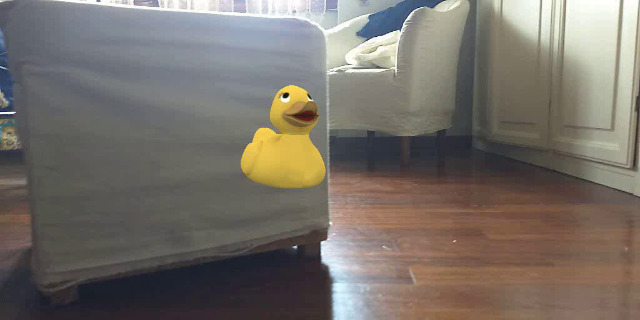} &
    \includegraphics[width=0.45\linewidth]{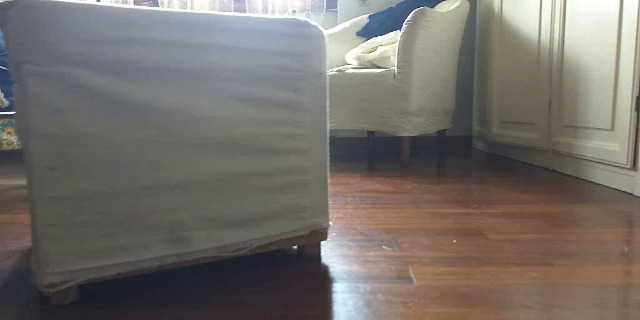} \\
    \includegraphics[width=0.45\linewidth]{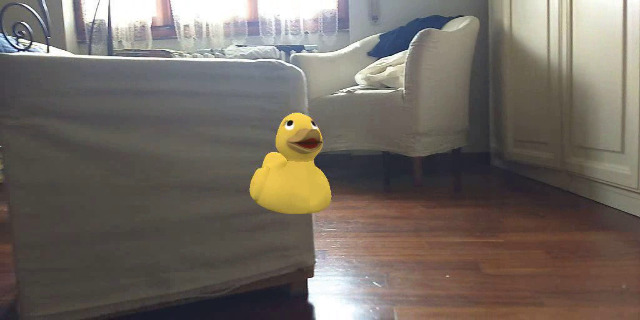} &
    \includegraphics[width=0.45\linewidth]{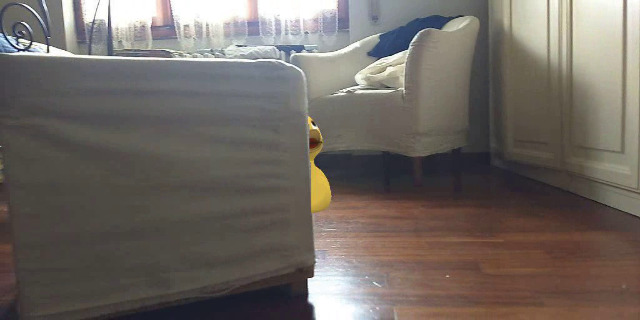} \\
    AR w/o O.H. & AR with our O.H. \\
    \end{tabular}

    \caption{\textbf{AR with occlusion handling (O.H.).} On the left, vanilla AR enabled by an Android device with ARCore. On the right, instead, our depth-aware AR enabled by single image depth prediction with \pyd{} for occlusion handling.}
    \label{fig:occlusion_handling}
\end{figure}

Differently from other approaches, such as \cite{Occlusion2018} and \cite{Luo-VideoDepth-2020}, our networks do not require SLAM points to infer dense depth maps nor a fine-tuning of the network on the input video data. In our case, a single image and at least one point in scale suffice to obtain absolute depth perception. Consequently, we do not rely on other techniques (\eg{} optical flow or edge localization) in our whole pipeline for AR. Nevertheless, it can be noticed in Figure \ref{fig:ar_comparison} how our strategy coupled with \pyd{} can produce competitive and detailed depth maps leveraging a single RGB image only.
Figure \ref{fig:occlusion_handling} shows some qualitative examples of an AR application, \ie{} visualization of a \textit{virtual duck} in the observed scene. Once positioned on a surface, we can notice how foreground elements do not correctly hide it without proper occlusion handling. In contrast, our strategy allows for a more realistic experience, thanks to the dense and robust depth map inferred by \pyd{} and sparse anchors provided by the AR framework.

\section{Conclusion}
In this paper, we proposed a strategy to train single image depth estimation networks, focusing our attention on lightweight ones suited for handheld devices characterized by severe constraints concerning power consumption and computational resources. An exhaustive evaluation highlights that real-time depth estimation from a single image in the wild is feasible by adopting appropriate network design and training strategies.
By distilling knowledge from complex architecture, not suited for mobile deployment, we have shown that it is possible to develop accurate yet fast networks enabling for a variety of AR applications on consumer smartphones.
We also reported the effectiveness of such an approach in two notable application scenarios concerning depth aware blurring and augmented reality.

\section*{Acknowledgement}
We gratefully acknowledge the support of NVIDIA Corporation with the donation of the Titan Xp GPU used for this research.

\bibliographystyle{IEEEtran}
\bibliography{egbib}

\end{document}